# Exploiting Causality for Selective Belief Filtering in Dynamic Bayesian Networks

**Stefano V. Albrecht**                                    SVALB@CS.UTEXAS.EDU
*Department of Computer Science*
*The University of Texas at Austin*
*Austin, TX 78712, USA*

**Subramanian Ramamoorthy**                          S.RAMAMOORTHY@ED.AC.UK
*School of Informatics*
*The University of Edinburgh*
*Edinburgh, EH8 9AB, UK*

## Abstract

Dynamic Bayesian networks (DBNs) are a general model for stochastic processes with partially observed states. Belief filtering in DBNs is the task of inferring the belief state (i.e. the probability distribution over process states) based on incomplete and noisy observations. This can be a hard problem in complex processes with large state spaces. In this article, we explore the idea of accelerating the filtering task by automatically exploiting causality in the process. We consider a specific type of causal relation, called *passivity*, which pertains to how state variables cause changes in other variables. We present the *Passivity-based Selective Belief Filtering* (PSBF) method, which maintains a factored belief representation and exploits passivity to perform *selective updates* over the belief factors. PSBF produces exact belief states under certain assumptions and approximate belief states otherwise, where the approximation error is bounded by the degree of uncertainty in the process. We show empirically, in synthetic processes with varying sizes and degrees of passivity, that PSBF is faster than several alternative methods while achieving competitive accuracy. Furthermore, we demonstrate how passivity occurs naturally in a complex system such as a multi-robot warehouse, and how PSBF can exploit this to accelerate the filtering task.

## 1. Introduction

Dynamic Bayesian networks (DBNs) (Dean & Kanazawa, 1989) are a general model for stochastic processes with partially observed states. The topology of a DBN is a compact specification of how variables in the process interact during transitions (cf. Figure 1). Given the possible incompleteness and noise in observations, it may not generally be possible to infer the state of the process with absolute certainty. Instead, we may infer beliefs about the process state based on the history of observations, in the form of a probability distribution over the state space of the process. This is often called a *belief state* and the task of calculating belief states is commonly referred to as *belief filtering*.

A number of exact and approximate inference methods exist for Bayesian networks (see, e.g., Koller & Friedman, 2009; Pearl, 1988) which can be used for filtering in DBNs, by applying them to the "unrolled" DBN in which the $t + 1$ slice is repeated for each observed time step, or via a successive update in which the current posterior (belief state) is used as the prior in the next time step (see also Murphy, 2002). However, it is clear that the unrolled variant becomes intractable as the network grows unboundedly with time. Even in the successive update, exact methods become intractable in high-dimensional process states and approximate methods may propagate growing errors over time. Therefore, filtering methods were developed which utilise the special structure of DBNs and maintain the errors propagated over time. (We defer a detailed discussion of such methods to Section 2.)

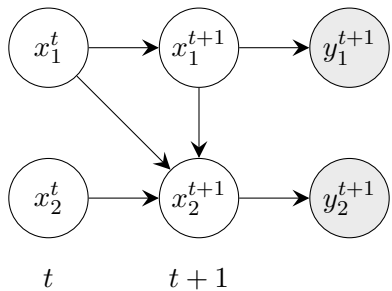

Figure 1: Example of a dynamic Bayesian network (DBN) with two state variables and two observation variables. The $x_i^t$ and $x_i^{t+1}$ variables represent the process states at time $t$ and $t + 1$, respectively, while the $y_i^{t+1}$ variables (shaded) represent the observation at time $t + 1$. The arrows describe how the variables interact.

Often, the key to developing efficient filtering methods is to identify structure in the process which can be leveraged for inference. In this article, we are interested in the application of DBNs as representations of actions in partially observed decision processes, such as POMDPs (Kaelbling, Littman, & Cassandra, 1998; Sondik, 1971) and their many variants. DBNs can be used to represent the effects of actions on the decision process, by specifying how variables interact and what information the decision maker observes. In many cases, decision processes exhibit high degrees of *causal structure* (Pearl, 2000), by which we mean that a change in one part of the process may cause a change in another part. Our experience with such processes is that this causal structure may be used to make the filtering task more tractable, because it can tell us that beliefs need only be revised for certain aspects of the process state. For example, if the variable $x_2$ in Figure 1 changes its value only if variable $x_1$ changed its value (i.e. a change in $x_1$ *causes* a change in $x_2$), then it seems intuitive to use this causal relation when deciding whether to revise one's belief about $x_2$. Unfortunately, current filtering methods do not take such causal structure into account.

We refer to the above type of causal relation (between $x_1$ and $x_2$) as *passivity*. Intuitively, we say that a state variable $x_i$ is passive in a given action if, when executing that action, there is a subset of the state variables that directly affect $x_i$ (i.e. $x_i$'s parents in the DBN) such that $x_i$ may change its value only if at least one of the variables in this subset changed its value. It is worth pointing out that passivity occurs naturally and frequently in many planning domains, especially in robotic and other physical systems (Mainzer, 2010). The following example[1] illustrates this in a simple robot arm:

---

1. We mark the end of an example with a solid black square.

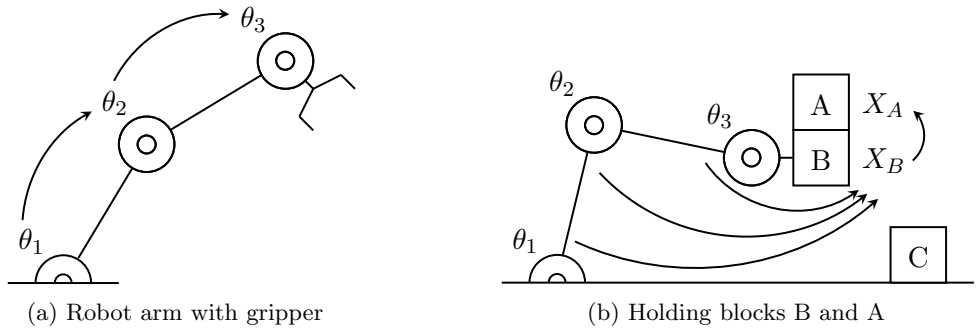

(a) Robot arm with gripper     (b) Holding blocks B and A

Figure 2: Robot arm with three rotational joints and gripper. The variables $\theta_i$ represent the absolute orientations of the corresponding joints.

**Example 1** (Robot arm). Consider a robot arm with three rotational joints and a gripper, as shown in Figure 2a. The joints are denoted by $\theta_1, \theta_2, \theta_3$ and may take any values from the discrete set $\{0°, 1°, ..., 359°\}$ which indicate their absolute orientations (e.g. $\theta_i = 0°$ means that joint $i$ points exactly to the right, $\theta_i = 180°$ means that it points to the left). For each joint $i$, let there be two actions $CW_i$ and $CCW_i$ which rotate the joint by $1°$ clockwise and counter-clockwise, respectively. The uncertainty in this system could be due to stochastic joint movements or unreliable sensor readings for the joint orientations.

For any action $CW_i$ or $CCW_i$, the variable $\theta_i$ is *not* passive because its value is directly modified by the action. However, the variables $\theta_{j \neq i}$ are passive because they change their values only if the corresponding preceding variable $\theta_{j-1}$ changed its value, since a changed orientation of joint $j - 1$ *causes* a changed orientation of joint $j$ (recall that the orientations are absolute). Note that this also accounts for *chains* of such causal effects, as indicated by the arrows: the orientation of joint 3 changes if the orientation of joint 1 changes, since joint 1 causes joint 2 to change, which in turn causes joint 3 to change.

Further examples of passivity can be seen in the context of object manipulation, such as in the "blocks" planning domain (e.g. Pasula, Zettlemoyer, & Kaelbling, 2007). Figure 2b shows the arm holding blocks B and A, with A on top of B. Here, the position of B ($X_B$) is passive with respect to the joint orientations since it will only change if any of the orientations changed. Furthermore, there is a causal chain from the joint orientations to the position of block A ($X_A$), since A's position will change if B's position changes. ∎

How can passivity be exploited to accelerate the filtering task in the above example? The fact that the state variables are passive means that some aspects of the state may remain unchanged, depending on which action we choose. For example, if we choose to rotate joint 3, then the fact that joints 1 and 2 are passive means that they are unaffected by this action. Thus, it seems redundant to revise beliefs for the orientations of joints 1 and 2. However, this is precisely what current filtering methods do (cf. Section 2).

More concretely, assume we use a factored belief representation $P(\theta_1, \theta_2, \theta_3) = P(\theta_1, \theta_2) * P(\theta_2, \theta_3)$ and choose to rotate $\theta_3$ in any direction. Then, it is easy to see that we will need to update the factor $P(\theta_2, \theta_3)$, since $\theta_3$ changes its value, but not the factor $P(\theta_1, \theta_2)$, since the variables $\theta_1, \theta_2$ are both passive. Since the parents of $\theta_1, \theta_2$ (if any) do not change their values, we know that $\theta_1, \theta_2$ will not change their values either. As we will show later, skipping over $P(\theta_1, \theta_2)$ does not result in a loss of information in such cases, and similarly for chains of such causal connections (cf. Example 1). A more complex example of a planning domain involving passivity, and how it can be exploited, is discussed in Section 6.2.

In addition to guiding belief revision, there are several features which make passivity an interesting example of a causal relation: First of all, passivity is a latent causal relation, meaning that it can be readily extracted from the process dynamics without additional annotation by an expert. (In Section 4, we give a procedure which identifies passive variables based on their conditional probability tables.) Furthermore, passivity is not a deterministic relation since passive variables may have any stochastic behaviour when changing their values. Finally, passivity is a relatively simple example of a causal relation, and the idea of exploiting passivity in order to accelerate the filtering task is intuitive. Yet, to the best of our knowledge, this has not been formalised and explored rigorously before.

The purpose of the present article is to formalise and evaluate the idea of *automatically exploiting* causal structure for efficient belief filtering in DBNs, using passivity as a concrete example of a causal relation. Specifically, our hypothesis is that in large processes with high degrees of passivity, this structure can be exploited to accelerate the filtering task. After discussing related work in Section 2 and technical preliminaries in Section 3, our contributions can be grouped into the following parts:

- In Section 4, we give a formally concise definition of passivity and discuss various aspects of this definition. Our definition assumes a decision process which is specified as a set of dynamic Bayesian networks (one for each action). We also discuss a non-example of passivity, by which we mean variables which appear to be passive but really are not passive. Finally, we give a simple procedure which can detect passive variables based on their conditional probability tables.

- In Section 5, we present the *Passivity-based Selective Belief Filtering* (PSBF) method. Following the idea outlined above, PSBF uses a factored belief representation in which the belief factors are defined over clusters of correlated state variables. PSBF follows a 2-step update procedure wherein the belief state is first propagated through the process dynamics (the *transition step*) and then conditioned on the observation (the *observation step*). The interesting novelty of PSBF is the way in which it performs the transition step: rather than updating all belief factors, PSBF updates only those factors whose variables it suspects to have changed, which is possible by exploiting passivity (to be made precise shortly). Similarly, in the observation step, PSBF updates only those belief factors which it determines to be structurally connected with the observation, and it uses only those parts of the observation which are relevant to the belief factor, thus allowing for a more efficient incorporation of observations. PSBF produces exact belief states under certain assumptions and approximate belief states otherwise. We also discuss the computational complexity and error bounds of PSBF.

- In Section 6, we evaluate PSBF in two experimental domains: We first evaluate PSBF in synthetic (i.e. randomly generated) processes of varying sizes and degrees of passivity. The process sizes vary from one thousand to one trillion states, and the passivity degrees vary from 25% to 100% passivity. Our results show that PSBF is faster than several alternative methods while maintaining competitive accuracy. In particular, our results indicate that the computational gains grow significantly with both the degree of passivity and the size of the process. We then evaluate PSBF in a complex simulation of a multi-robot warehouse system in the style of Kiva (Wurman, D'Andrea, & Mountz, 2008). We show how passivity occurs in this system and how PSBF can exploit this to accelerate the filtering task, again outperforming alternative methods.

Finally, we discuss the strengths and weaknesses of PSBF in Section 7, and we conclude our work in Section 8. All proofs can be found in the appendix.

## 2. Related Work

There exists a substantial body of work on belief filtering in partially observed stochastic processes. In this section, we review filtering methods that utilise the special structure of DBNs and situate our work within this and other related literature.

### 2.1 Approximate Belief Filtering in DBNs

Several authors proposed filtering methods wherein the belief state is represented as a set of state samples. Specifically, the probability that the process is in state $s$ is the normalised frequency with which the state samples correspond to $s$. These methods are now commonly referred to as *particle filters* (PF); see the work of Doucet, de Freitas, and Gordon (2001) for a survey. In a common variant of PF (Gordon, Salmond, & Smith, 1993), the filtering task consists of propagating the current state samples through the process dynamics and a subsequent resampling step based on the probabilities with which the new state samples would have produced the observation. Two interesting features of PF are that it can be applied to processes with discrete and continuous variables, and that the approximation error converges to zero as we increase the number of state samples.

A known problem of PF is the fact that the number of samples needed for acceptable approximations can grow drastically with the variance in the process dynamics (as shown in our experiments; cf. Section 6). Rao-Blackwellised PF (RBPF) (Doucet, De Freitas, Murphy, & Russell, 2000) was developed to address this problem. RBPF assumes that the state variables can be grouped into sets $R$ and $X$ such that the distribution over $X$ can be efficiently calculated from $R$ during the filtering. Hence, a sample in RBPF consists of a sample of $R$ and a corresponding marginal distribution over $X$. RBPF is useful when the variance in $R$ is relatively low and the variance in $X$ is high, since this reduces the number of samples needed for acceptable approximations.

Boyen and Koller (1999, 1998) recognised that if a process consists of several independent or weakly interacting subcomponents, then the belief state can be represented more efficiently as a product of smaller beliefs about these individual subcomponents. Their seminal contribution is to show that the approximation error due to this factored representation is essentially bounded by the degree of uncertainty (or "mixing rates") in the process. More precisely, they prove that the relative entropy (or KL divergence; Kullback & Leibler, 1951) between two belief states contracts at an exponential rate when propagated through a stochastic transition process. Based on this observation, they propose a filtering method (BK) wherein the belief state is represented in factored form and the belief factors are updated using an exact inference method, such as the junction tree algorithm (Lauritzen & Spiegelhalter, 1988). Since the internal "cliques" used in the junction tree algorithm may not correspond to the belief state representation of BK, a final "projection step" will typically have to be performed in which the original factorisation is restored. The performance of this method depends crucially on whether the relevant correlations between state variables can be captured in small clusters, and whether the projection step can be performed efficiently.

*Factored particle filtering* (FP) (Ng, Peshkin, & Pfeffer, 2002) addresses the main drawbacks of PF (many samples needed) and BK (small clusters required) by approximating the belief factors using a set of factored state samples. The samples are factored in the sense that they only assign values to the variables in the corresponding factor. This allows FP to represent belief factors which are too large for BK, and it reduces the number of samples needed due to the smaller number of variables in each factor. The authors provide different methods of updating the factored state samples, but the generic idea is to first perform a "join" operation in which full state samples are reconstructed from the factored samples, which are then updated as in standard PF. The updated samples are then projected down into factored form using a "project" operation. The main drawback of FP is that these join and project operations essentially correspond to standard relational database operations, which can be very expensive.

Murphy and Weiss (2001) propose a filtering method called *factored frontier* (FF). FF uses a fully factored representation of belief states; that is, the belief state is a product of marginals for each individual state variable. This allows for a very compact representation of beliefs. The algorithm works by "moving" a set of state variables (the frontier) forward and backward in the DBN topology. This requires a certain variable ordering, which can be difficult to attain if intra-correlations between state variables (i.e. edges within the $t + 1$ slice of the DBN) are allowed. The authors show that their method is equivalent to a single iteration of loopy belief propagation (LBP) (Pearl, 1988). Thus, similar to LBP, FF can be applied in successive iterations to improve the approximation accuracy.

None of the works discussed above explicitly address the question of how causal relations between state variables can be exploited to accelerate the filtering task, or, alternatively, how the filtering methods proposed therein implicitly benefit from causal structure. Our method, PSBF, is related to BK and FP in that PSBF, too, uses a factored belief representation, where the belief factors are defined over clusters of correlated state variables. Therefore, the analysis of approximation errors by Boyen and Koller (1998) also applies to PSBF, as we show in Section 5 as well as in our experiments. However, in contrast to BK and FP, PSBF does not perform inference over the complete factorisation, but rather over the individual factors. As a consequence, PSBF does not require a join or project operation, which is one of the main disadvantages of BK and FP.

## 2.2 Belief Filtering in Decision Processes

The methods discussed in the preceding subsection can be used for belief filtering in decision processes, including POMDPs (Kaelbling et al., 1998; Sondik, 1971). In this regard, these methods can be viewed as "pure" filters in that they are only concerned with belief filtering and not with the control of the decision process. This is in contrast to *combined* filtering methods, which interleave the filtering and control tasks in decision processes and make specific assumptions regarding solutions thereof. There exists a large body of literature on such combined methods, including reachability-based methods (Hauskrecht, 2000; Washington, 1997), grid-based methods (Zhou & Hansen, 2001; Brafman, 1997; Lovejoy, 1991), point-based methods (Smith & Simmons, 2005; Pineau, Gordon, & Thrun, 2003), and compression methods (Roy, Gordon, & Thrun, 2005; Poupart & Boutilier, 2002).

A potential advantage of such combined methods is that they have access to additional structure and may, therefore, utilise synergies between the filtering and control tasks. One such synergy is the use of decision quality to guide belief filtering, rather than metrics such as relative entropy. Poupart and Boutilier (2001, 2000) propose a filtering method, called *value-directed approximation*, which chooses different approximation schemes for different decisions so as to minimise the expected loss in decision quality (i.e. accumulated rewards). The method assumes that the POMDP has been solved exactly and that the value function is provided in the form of $\alpha$-vectors which represent the available actions in the POMDP. Based on the value function, their algorithm computes a "switching set" and "alternative plans" to determine the error bounds of approximation schemes. This is used to search for an optimal approximation scheme in a tree-based manner, where the search traverses from approximate to exact schemes.

While the idea of using decision quality to guide belief filtering is appealing, their method involves a series of optimisation problems and an exhaustive tree search, which can be very costly in complex systems. The advantage of pure filtering methods, including our proposed method PSBF, is that they can filter processes which are too complex for combined methods, such as the multi-robot warehouse system studied in Section 6. The actual control task can then be done via domain-specific solutions (cf. Section 6.2.1).

### 2.3 Substructure in Parameterisation

Bayesian networks, and hence DBNs, allow for a compact parameterisation (i.e. specification of probabilities) and efficient inference via conditional independence relations. In addition, there has been considerable work in identifying substructure in the parameterisation to further simplify knowledge acquisition and enhance inference (Koller & Friedman, 2009; Boutilier, Dean, & Hanks, 1999). The property studied in this work, passivity, is one example of substructure in the parameterisation. Other notable examples include causal independence (e.g. Heckerman & Breese, 1994; Heckerman, 1993) and context-specific independence (Boutilier, Friedman, Goldszmidt, & Koller, 1996).

Causal independence is the assumption that the effects of individual causes on a common variable (i.e. the parents of that variable) are independent of one another. This allows for a compact parameterisation via operators such as "noisy-or" (Srinivas, 1993; Pearl, 1988), and it can be used to enhance inference (Zhang & Poole, 1996). Note that passivity is a conceptually much simpler property than causal independence, because passivity is neither concerned with the strength of individual causes nor the extent to which they depend on each other. Moreover, passivity can be read directly from the parameterisation (cf. Section 4.3) whereas causal independence is usually imposed by the designer.

Context-specific independence (CSI) is a property which states that a variable is independent of some of its parents given a certain assignment of values (i.e. "context") to some of its other parents. Non-local CSI statements follow similarly to d-separation (Geiger, Verma, & Pearl, 1989). This can allow for a further reduction of parameters (Boutilier et al., 1996) and enhancement of inference (Poole & Zhang, 2003). As we will discuss in Section 4, passivity can be viewed as a special kind of CSI applied to DBNs, in that the parents with respect to which the variable is passive provide the context for CSI. However, in contrast to CSI, passivity does not assume that the context is actually observed.

## 3. Technical Preliminaries

This section introduces the basic concepts and notation used in our work. We begin with a brief discussion of decision processes to provide the context for our work, followed by a discussion of dynamic Bayesian networks as the model over which we perform inference.

### 3.1 Decision Processes, Belief States, Exact Updates

We consider a stochastic decision process wherein, at each time $t$, the process is in state $s^t \in S$ and a decision maker, or "agent", is choosing an action $a^t$. After executing $a^t$ in $s^t$, the process transitions into state $s^{t+1} \in S$ with probability $T^{a^t}(s^t, s^{t+1})$ and the agent receives an observation $o^{t+1} \in O$ with probability $\Omega^{a^t}(s^{t+1}, o^{t+1})$. We assume factored representations of the state space $S$ and observation space $O$, such that $S = X_1 \times ... \times X_n$ and $O = Y_1 \times ... \times Y_m$, where the domains $X_i, Y_j$ are finite. The notation $s_i$ is used to denote the value of $X_i$ in state $s \in S$, and analogously for $o_j$ with $o \in O$. Moreover, we assume that the process is time-invariant, meaning that $T^a$ and $\Omega^a$ are independent of $t$. This framework is compatible with many decision models used in the artificial intelligence literature, including POMDPs (Kaelbling et al., 1998; Sondik, 1971) and its many variants.

The agent chooses action $a^t$ based on its *belief state* $b^t$ (also known as *information state*), which represents the agent's beliefs about the likelihood of states at time $t$. Formally, a belief state is a probability distribution over the state space $S$ of the process. Belief filtering is the task of calculating a belief state based on the history of observations. Ideally, the resulting belief state should be *exact* in that it retains all relevant information from the past observations (this is sometimes referred to as *sufficient statistic*; cf. Astrom, 1965). The exact update rule is a simple procedure that produces exact belief states:

**Definition 1** (Exact update rule)**.** The *exact update rule* is defined as follows: After taking action $a^t$ and observing $o^{t+1}$, the belief state $b^t$ is updated to $b^{t+1}$ via

$$\hat{b}^{t+1}(s') \quad = \quad \sum_{s \in S} b^t(s) \, T^{a^t}(s, s') \tag{1}$$

$$b^{t+1}(s') \quad = \quad \eta \, \hat{b}^{t+1}(s') \, \Omega^{a^t}(s', o^{t+1}) \tag{2}$$

where $\eta$ is a normalisation constant.

We sometimes refer to the step $b^t \to \hat{b}^{t+1}$ as the *transition step* and to the step $\hat{b}^{t+1} \to b^{t+1}$ as the *observation step*. Unfortunately, the space complexity of storing exact belief states and the time complexity of updating them using the exact update rule are both exponential in the number of state variables, making it infeasible for complex systems with large state spaces. Hence, more efficient approximate methods are required.

### 3.2 Dynamic Bayesian Networks

A dynamic Bayesian network (DBN) (Dean & Kanazawa, 1989) is a Bayesian network with a special temporal semantics that specifies how a stochastic process transitions from one state into another. DBNs can be used to model the effects of actions in a stochastic decision process. Specifically, they are a compact representation of the transition function $T^a$ and observation function $O^a$ of action $a$:

**Definition 2** (DBN). A *dynamic Bayesian network* for action $a$, denoted $\Delta^a$, is an acyclic directed graph consisting of:

- State variables $X^t = \{x_1^t, ..., x_n^t\}$ and $X^{t+1} = \{x_1^{t+1}, ..., x_n^{t+1}\}$ with $x_i^t, x_i^{t+1} \in X_i$, representing the states of the process at time $t$ and $t + 1$, respectively.

- Observation variables $Y^{t+1} = \{y_1^{t+1}, ..., y_m^{t+1}\}$ with $y_j^{t+1} \in Y_j$, representing the observation received at time $t + 1$.

- Directed edges $E_a \subseteq (X^t \times X^{t+1}) \cup (X^{t+1} \times X^{t+1}) \cup (X^{t+1} \times Y^{t+1}) \cup (Y^{t+1} \times Y^{t+1})$, specifying the network topology and dependencies between variables.

- Conditional probability distributions $P_a(z \,|\, pa_a(z))$ for each variable $z \in X^{t+1} \cup Y^{t+1}$, specifying the probability that $z$ assumes a certain value given a specific assignment to its parents $pa_a(z) = \{z' \,|\, (z', z) \in E_a\}$. For convenience, we also define $pa_a^t(Z) = X^t \cap pa_a(Z)$ and $pa_a^{t+1}(Z) = X^{t+1} \cap pa_a(Z)$, where $pa_a(Z) = \cup_{z \in Z} pa_a(Z)$.

The edges $E_a$ and distributions $P_a$ define the functions $T^a$ and $\Omega^a$ as

$$T^a(s, s') \;=\; \prod_{i=1}^{n} P_a \left( x_i^{t+1} = s_i' \,|\, pa_a(x_i^{t+1}) \leftarrow (s, s') \right) \tag{3}$$

$$\Omega^a(s', o) \;=\; \prod_{j=1}^{m} P_a \left( y_j^{t+1} = o_j \,|\, pa_a(y_j^{t+1}) \leftarrow (s', o) \right) \tag{4}$$

where we use the notation $pa_a(x_i^{t+1}) \leftarrow (s, s')$ to specify that the parents of $x_i^{t+1}$ in $X^t$ and $X^{t+1}$, respectively, assume their corresponding values from $s$ and $s'$. Formally, if $x_l^t \in pa_a^t(x_i^{t+1})$ and $x_{l'}^{t+1} \in pa_a^{t+1}(x_i^{t+1})$, then $x_l^t = s_l$ and $x_{l'}^{t+1} = s_{l'}'$. Similarly, we use the notation $pa_a(y_j^{t+1}) \leftarrow (s', o)$ to specify that the parents of $y_j^{t+1}$ in $X^{t+1}$ and $Y^{t+1}$, respectively, assume corresponding values from $s'$ and $o$.

**Example 2** (DBN representation of robot arm). We can represent the robot arm from Example 1 as a set of DBNs, where we have one DBN $\Delta^a$ for each action $a \in \{CW_i, CCW_i\}$. The state and observation variables in the DBNs are $X^t = \{\theta_1^t, \theta_2^t, \theta_3^t\}$, $X^{t+1} = \{\theta_1^{t+1}, \theta_2^{t+1}, \theta_3^{t+1}\}$, and $Y^{t+1} = \{\hat{\theta}_1^{t+1}, \hat{\theta}_2^{t+1}, \hat{\theta}_3^{t+1}\}$. To make our example more realistic, let us assume that the joint orientations are bounded relative to the orientation of the immediately preceding joint (e.g. in the form of a cone), where the first joint is bounded relative to the ground. This means that the joint movement depends on its own as well as the preceding joint orientation, as shown in Figure 3. Moreover, the joint orientations are correlated (i.e. edges within $X^{t+1}$) such that no joint can exceed the bound given by the preceding joint. Finally, the observation variables depend solely on the corresponding joint variable. The actions in this example would differ in their variable distributions $P_a$. ∎

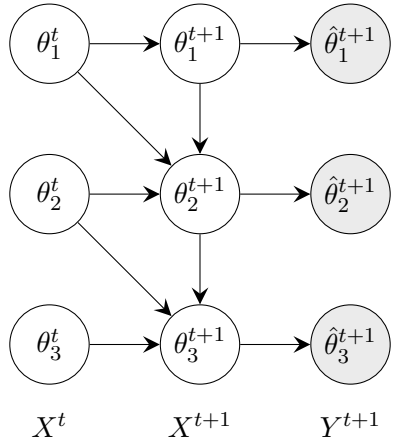

Figure 3: DBN representation of robot arm.

## 3.3 Additional Definitions

It will be useful to define the following:

- The binary order $\prec$ is defined over $X^t \cup X^{t+1}$ such that $x_i^t \prec x_j^t$ and $x_i^{t+1} \prec x_j^{t+1}$ for all $1 \le i < j \le n$, and $x_i^t \prec x_j^{t+1}$ for all $1 \le i, j \le n$.

- Given a set $Z \subseteq X^t \cup X^{t+1}$, we write $Z^{\prec}$ to denote the tuple that contains all variables of $Z$, ordered by $\prec$.

- Given the ordered tuple $Z^{\prec} = (z_{i_1}, ..., z_{i_{|Z|}})$, we define the set $S(Z) = X_{i_1} \times .... \times X_{i_{|Z|}}$ to contain all value tuples for the variables in $Z$.

- Given a value tuple $s_Z = (s_{i_1}, ..., s_{i_{|Z|}}) \in S(Z)$, we use the notation $Z \hookleftarrow s_Z$ as an abbreviation for $z_{i_l} = s_{i_l}$ for each $z_{i_l} \in Z^{\prec}$ (i.e. the variables in $Z$ assume their corresponding values from $s_Z$).

## 4. Passivity

This section introduces a formal definition of passivity, which will then be used as the basis for the remainder of this article. We also provide a simple procedure to detect passive variables from the process dynamics.

### 4.1 Formal Definition

As outlined in Section 1, a state variable $x_i^{t+1}$ is called passive in action $a$ if there exists a subset of $x_i^{t+1}$'s parents in $X^t$ (in the DBN $\Delta^a$) such that $x_i^{t+1}$ may change its value only if at least one of the variables in this subset changed its value. Conversely, $x_i^{t+1}$ does not change if the variables in the subset did not change. Formally, we define passivity as follows:

**Definition 3** (Passivity). Let action $a$ be given by a DBN $\Delta^a$. A state variable $x_i^{t+1}$ is called *passive* in $\Delta^a$ if there exists a set $\Phi_{a,i} \subseteq pa_a^t(x_i^{t+1}) \setminus \{x_i^t\}$ such that:

(i) $\forall x_j^t \in \Phi_{a,i} : \left(x_j^{t+1}, x_i^{t+1}\right) \in E_a$
and

(ii) for any two states $s^t$ and $s^{t+1}$ with $T^a(s^t, s^{t+1}) > 0$ :

$$\left(\forall x_j^t \in \Phi_{a,i} : s_j^t = s_j^{t+1}\right) \Rightarrow s_i^t = s_i^{t+1} \tag{5}$$

A state variable which is not passive is called *active*.

The set $\Phi_{a,i}$ corresponds to the subset of variables described above: it contains all those variables which directly affect $x_i^{t+1}$ (i.e. they are parents of $x_i^{t+1}$ in $X^t$) such that $x_i^{t+1}$ may change its value only if any of the variables in $\Phi_{a,i}$ changed its value. We will sometimes say that a variable $x_i^{t+1}$ is passive in $\Delta^a$ *with respect to* another variable $x_j^t$ if it is the case that $x_j^t \in \Phi_{a,i}$. Furthermore, we will omit "in $\Delta^a$" if it is obvious from context.

Clause (i) in Definition 3 requires that $x_i^{t+1}$ is intra-correlated with the variables in $\Phi_{a,i}$; specifically, that there is an edge from $x_j^{t+1}$ to $x_i^{t+1}$ for all $x_j^t \in \Phi_{a,i}$. As an example, see Figure 1 in which we assumed that the variable $x_2^{t+1}$ was passive with respect to the variable $x_1^t$. (We will discuss the purpose of this clause in the next subsection.) Clause (ii) defines the core semantics of passivity by requiring that $x_i^{t+1}$ remains unchanged if all variables in $\Phi_{a,i}$ remain unchanged. Note that this means that the distribution $P_a$ for $x_i^{t+1}$ may specify any deterministic or stochastic behaviour if the variables in $\Phi_{a,i}$ change their values. This includes that $x_i^{t+1}$ may not change its value at all.

A state variable $x_i^{t+1}$ can be passive even if it has no parents in $X^t$, or none other than $x_i^t$. In this case, the set $\Phi_{a,i}$ would be empty and clause (i) as well as the premise in (5) would trivially hold true. However, such a variable can only be passive if it does not change its value under any circumstances. In other words, it would have to be a constant. In that case, one should consider removing the variable from the state description in order to reduce computational costs.

As noted in Section 2.3, passivity can be shown to be a special kind of context-specific independence (CSI) (Boutilier et al., 1996) applied to DBNs. Here, the associated set $\Phi_{a,i}$ of a passive variable $x_i^{t+1}$ provides the context: given any assignment of values to $x_j^t \in \Phi_{a,i}$ (i.e. context) such that $x_j^t = x_j^{t+1}$, $x_i^{t+1}$ is independent of all $x_k^t, x_k^{t+1}$ with $x_k^t \in pa_a^t(x_i^{t+1}) \setminus \Phi_{a,i}$ and $k \neq i$. However, besides this similarity, there is an important difference between passivity and CSI, which is that passivity does not actually assume that the context is observed. Thus, passivity can be viewed as a kind of CSI for *unobserved* contexts. This will become clear in Section 5, when we describe a filtering method that exploits passivity.

## 4.2 Non-Example of Passivity

What is the purpose of clause (i) in the definition of passivity? After all, and as discussed previously, clause (ii) captures the core idea of passivity, which is that a variable may only change its value if any of the variables with respect to which it is passive changed its value.

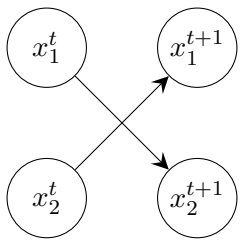

Figure 4: Example of a process for which clause (ii) is insufficient.

However, while it may seem intuitive that clause (ii) be sufficient for passivity, there are in fact processes in which clause (ii) alone does not suffice. In other words, clause (ii) is necessary but not sufficient for passivity. We illustrate this in the following example:

**Example 3** (Non-example of passivity). Consider a process with two binary state variables, $x_1, x_2$, and a single action, $a$, shown in Figure 4. (We omit the observation variables for clarity.) The dynamics of the process are such that $x_1^{t+1}$ takes the value of $x_2^t$ and $x_2^{t+1}$ takes the value of $x_1^t$ (i.e. $x_1$ and $x_2$ swap their values at each time step). In this process, both state variables satisfy clause (ii) of Definition 3: If we set $x_1^0 = x_2^0$ (i.e. same initial values), then $T^a(s^t, s^{t+1})$ is positive only for states $s^t = s^{t+1}$, and hence (5) is true. If we set $x_1^0 \neq x_2^0$, then $T^a(s^t, s^{t+1})$ is positive only for states $s^t, s^{t+1}$ with $s_i^t \neq s_i^{t+1}, i \in \{1, 2\}$, and hence (5) is trivially true since its premise is false. ∎

Despite satisfying clause (ii), the state variables $x_1^{t+1}$ and $x_2^{t+1}$ from Example 3 are in fact not passive, for the following two reasons: Firstly, passivity is a causal relation and as such it must imply a *causal order* (Pearl, 2000). However, there is no causal order between $x_1$ and $x_2$, because there is no edge between $x_1^{t+1}$ and $x_2^{t+1}$. Secondly, passivity means that a variable may change its value only if another variable with respect to which it is passive (a variable in $\Phi_{a,i}$) changed its value. In other words, whether or not a passive variable $x_i^{t+1}$ may change its value depends on both the past values of $\Phi_{a,i}$ (at time $t$) *and* the new values of $\Phi_{a,i}$ (at time $t + 1$). However, the variables in Example 3 only depend on the values at time $t$, hence their own values at time $t + 1$ are predetermined and do not depend on whether the variables in $\Phi_{a,i}$ change values.

The first issue, namely that of the causal order, can be addressed by adding the corresponding edges in $X^{t+1}$. For instance, in Example 3 we could add an edge from $x_1^{t+1}$ to $x_2^{t+1}$ to establish a causal order. However, this does not generally solve the second issue, which is that every passive variable $x_i^{t+1}$ must depend on both past and new values of the variables in $\Phi_{a,i}$. In other words, $x_i^{t+1}$ must be both inter-correlated as well as intra-correlated with the variables in $\Phi_{a,i}$. The former is given by definition (since every variable in $\Phi_{a,i}$ is a parent of $x_i^{t+1}$) and the latter is precisely what is required by clause (i) in Definition 3. Therefore, clauses (i) and (ii) together define the formal meaning of passivity.

### 4.3 Detecting Passive Variables

As mentioned in Section 1, passivity is a *latent* causal property in the sense that it can be extracted from the process dynamics without additional information, and with no additional assumptions regarding the representation of variable distributions. In order to determine if a

---

**Algorithm 1** PASSIVE($x_i^{t+1}, \Delta^a$)

---

1: **Input:** state variable $x_i^{t+1}$, DBN $\Delta^a$

2: **Output:** $\Phi_{a,i}$ if $x_i^{t+1}$ is passive in $\Delta^a$, else **false**

3: $Q \leftarrow$ ORDEREDQUEUE $\left( \mathcal{P}\big(pa_a^t(x_i^{t+1}) \setminus \{x_i^t\}\big) \right)$ // *in ascending order of* $|\Phi_{a,i}|$

4: **while** $Q \neq \emptyset$ **do**

5:     $\Phi_{a,i} \leftarrow$ NEXTELEMENT($Q$)

6:     $Q \leftarrow Q \setminus \{\Phi_{a,i}\}$

7:     **for all** $x_j^t \in \Phi_{a,i}$ **do**

8:         **if** $\left(x_j^{t+1}, x_i^{t+1}\right) \notin E_a$ **then**

9:             Go to line 4 // *clause (i) violated*

10:     $\Psi_{a,i} \leftarrow pa_a(x_i^{t+1}) \setminus \big(\Phi_{a,i} \cup \{x_i^t\}\big)$

11:     $\Phi_{a,i}^{t+1} \leftarrow \left\{ x_j^{t+1} \mid x_j^t \in \Phi_{a,i} \right\}$

12:     **for all** $s_\Psi \in S(\Psi_{a,i}),\ s_\Phi \in S(\Phi_{a,i}),\ s_i \in X_i$ **do**

13:         **if** $P_a\left(x_i^{t+1} = s_i \mid x_i^t = s_i,\ \Phi_{a,i} \hookleftarrow s_\Phi,\ \Phi_{a,i}^{t+1} \hookleftarrow s_\Phi,\ \Psi_{a,i} \hookleftarrow s_\Psi\right) < 1$ **then**

14:             Go to line 4 // *clause (ii) violated*

15:     **return** $\Phi_{a,i}$

16: **return false**

---

variable $x_i^{t+1}$ is passive in $\Delta^a$, one has to find a set $\Phi_{a,i}$ such that both clauses of Definition 3 are satisfied. A simple procedure which does this for any representation of the variable distributions is given in Algorithm 1. The algorithm takes as inputs a variable $x_i^{t+1}$ and a DBN $\Delta^a$, and checks whether $x_i^{t+1}$ is passive in $\Delta^a$ by searching for a set $\Phi_{a,i}$ which satisfies both clauses of Definition 3. Note that the power set $\mathcal{P}$ in line 3 includes the empty set $\emptyset$, hence it also accounts for $\Phi_{a,i} = \emptyset$. Lines 7 to 9 check if clause (i) is satisfied while lines 10 to 14 check if clause (ii) is satisfied. Line 13 essentially checks if (5) holds true. If both clauses are satisfied, then $x_i^{t+1}$ is passive in $\Delta^a$ with respect to the variables in $\Phi_{a,i}$, and the algorithm returns the set $\Phi_{a,i}$. Otherwise, the algorithm returns a logical false.[2]

The time complexity of Algorithm 1 is exponential in the worst case, in which $x_i^{t+1}$ is not passive. Specifically, the time requirements of line 4 grow exponentially with the number of parents of $x_i^{t+1}$ in $X^t$, and the time requirements of line 12 grow exponentially with the cardinality of $\Phi_{a,i}$ and $\Psi_{a,i}$. However, these time requirements can be reduced significantly when committing to specific representations for the variable distributions $P_a$. For example, if the distributions are represented in tabular form, then one can utilise arrays of indices to perform sweeping tests of (5), i.e. line 13. Moreover, it is important to realise that the algorithm needs to be performed only once for each state variable, prior to the start of the process or on demand. This is since passivity is invariant of the process states. In other words, if a variable is passive in $\Delta^a$, then it will always be passive in $\Delta^a$. Therefore, it suffices to check once in advance for passivity.

---

2. Strictly speaking, Algorithm 1 checks for a property which is stronger than passivity because it does not check for $T^a(s^t, s^{t+1}) > 0$ (cf. clause (ii)) in line 12. However, the algorithm can be modified to include such a check. We omit this in our exposition in order to highlight the core ideas behind the algorithm.

Note that the set $\Phi_{a,i}$ is not necessarily unique. For example, consider a variable $x_1^{t+1}$ which is passive in $\Delta^a$ with respect to variables $x_2^t$ and $x_3^t$, i.e. $\Phi_{a,1} = \{x_2^t, x_3^t\}$, and assume that $x_2^{t+1}$ changes if and only if $x_3^{t+1}$ changes (i.e. they change at the same time). Then, it is easy to verify that $\Phi'_{a,1} = \{x_2^t\}$ and $\Phi''_{a,1} = \{x_3^t\}$ also satisfy clauses (i) and (ii), and hence $\Phi_{a,1}, \Phi'_{a,1}, \Phi''_{a,1}$ are all valid sets under our definition of passivity. The guiding principle in such cases is *Occam's razor*, which, intuitively speaking, states that the simplest explanation suffices. In our case, this means that it suffices to use the *smallest* set $\Phi_{a,i}$ in terms of the cardinality $|\Phi_{a,i}|$. (Hence, line 3 in Algorithm 1 sorts the queue $Q$ in ascending order of $|\Phi_{a,i}|$.) The rationale is that if there exist multiple causal explanations for a passive variable $x_i^{t+1}$, then the one involving the fewest key variables is to be favoured since it reduces (compared to the alternative explanations) the number of cases in which we would have to revise our beliefs about $x_i^{t+1}$. In our earlier example, if we accept $\Phi_{a,1}$ as a causal explanation for $x_1^{t+1}$, then we would have to revise our beliefs for $x_1^{t+1}$ every time $x_2^{t+1}$ or $x_3^{t+1}$ may have changed their values. However, if we accept $\Phi'_{a,1}$ as a causal explanation, then we would have to revise our belief for $x_1^{t+1}$ only if $x_2^{t+1}$ may have changed its value. This difference will become more obvious in Section 5.2, which explains how passivity can be exploited to reduce computational costs.

## 5. Passivity-based Selective Belief Filtering

This section presents the *Passivity-based Selective Belief Filtering* (PSBF) method, which exploits passivity for efficient filtering. As discussed in Section 3, we assume that the process is specified as a set of dynamic Bayesian networks which contains one DBN $\Delta^a$ for each action $a \in A$. Therefore, whenever we refer to an action $a$ (e.g. $T^a$, $\Omega^a$, $P_a$, $pa_a$), this is assumed to be in the context of $\Delta^a$.

PSBF follows the general two-step update procedure in which the belief state is first propagated through the process dynamics (*transition step*) and then conditioned on the observation (*observation step*). Thus, it is natural to divide the exposition of PSBF into three parts: (1) the belief state representation, (2) the transition step, and (3) the observation step. These are discussed in Sections 5.1, 5.2, and 5.3, respectively. A summary of PSBF is given in Section 5.4. We also discuss the computational complexity and error bounds of PSBF in Sections 5.5 and 5.6, respectively.

### 5.1 Belief State Representation

Recall from Section 1 that the principal idea behind PSBF is to maintain separate beliefs about individual aspects of the process, and to exploit passivity in order to perform selective updates over these separate beliefs. The union of all individual aspects constitutes a complete state description of the process. Therefore, the belief state can be represented as the product of all separate beliefs about the individual aspects.

We capture the informal notion of "individual aspects" formally in the form of *clusters*, which are defined as follows:

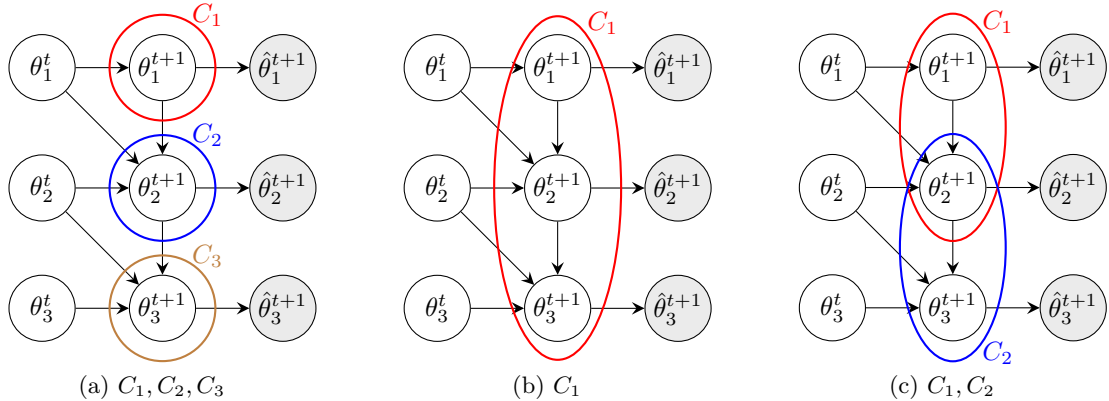

Figure 5: Three clusterings for the robot arm DBN.

**Definition 4** (Cluster). A *clustering* of $X^{t+1}$ is a set $\mathbb{C} = \{C_1, ..., C_K\}$ which satisfies $\forall k : C_k \subseteq X^{t+1}$ and $C_1 \cup ... \cup C_K = X^{t+1}$. We refer to the elements $C_k \in \mathbb{C}$ as *clusters*.

The underlying idea behind the concept of clusters is that the variables in a cluster $C_k$ are connected in some important sense. Specifically, if two or more variables are in a common cluster, then there exists some relation between these variables regarding the likelihood of values which they may assume. In other words, the variables are correlated in $X^{t+1}$.

The number $K$ and the concrete choice of clusters $C_k$ can be specified by the user or generated automatically. For example, they may be specified manually by a domain expert who is familiar with the structure of the modelled system, or generated automatically using methods such as the ones described in Section 6.1. It should be stressed, however, that in order to reduce computational costs, it is advisable to follow the general rule "as small as possible, as large as necessary" when choosing clusters (see Section 5.5 for a discussion about computational complexity). Therefore, if two variables are strongly correlated, then they should presumably be in a common cluster, whereas if they are not or only weakly correlated ("weakly" meaning that the correlation can be ignored safely), then they should be in separate clusters in order to reduce computational costs. This is illustrated in the following example:

**Example 4** (Clusters in robot DBN). Recall the robot arm DBN from Example 2, specifically Figure 3. One way to cluster the state variables in $X^{t+1}$ is given by the three clusters $C_1 = \{\theta_1^{t+1}\}$, $C_2 = \{\theta_2^{t+1}\}$, $C_3 = \{\theta_3^{t+1}\}$, as shown in Figure 5a. This clustering is most efficient since it minimises the size of each cluster. However, the clusters fail to capture the important correlation that the joint orientation $\theta_i$ is restricted by the preceding joint orientation $\theta_{i-1}$. Another way to cluster the state variables is given by the single cluster $C_1 = \{\theta_1^{t+1}, \theta_2^{t+1}, \theta_3^{t+1}\}$, as shown in Figure 5b. This clustering captures all correlations between variables. However, this is the largest possible cluster and, therefore, the least efficient one. A compromise is given by the two clusters $C_1 = \{\theta_1^{t+1}, \theta_2^{t+1}\}$, $C_2 = \{\theta_2^{t+1}, \theta_3^{t+1}\}$, which are shown in Figure 5c. This clustering captures the correlation of the joint orientations with the immediately preceding joint orientations, and it is more efficient than the previous clustering since it has smaller clusters. ∎

Given the definition of clusters, we capture the informal notion of "separate beliefs" in the form of *belief factors*:

**Definition 5** (Belief factor). Given a cluster $C_k$, the corresponding *belief factor* $b_k$ is a probability distribution over the set $S(C_k)$.

Intuitively, a belief factor $b_k$ represents the agent's beliefs as to the likelihood of values for the variables in the corresponding cluster $C_k$. An analogy to this is to view a belief factor as a "smaller" belief state, and to view $b$ as the "full" belief state which is a combination of the smaller belief states. However, to distinguish the two, we refer to $b$ simply as the belief state and to $b_k$ as a belief factor.

Finally, given the clusters $C_k$ and their corresponding belief factors $b_k$, the belief state $b$ is represented in factored form as

$$b(s) = \prod_{k=1}^{K} b_k(s_k)$$

where we use the notation $s_k$ to refer to the tuple $(s_i)_{x_i^{t+1} \in C_k}$. (E.g., if $C_k = \{x_2^{t+1}, x_3^{t+1}\}$ and $s = (s_1, s_2, s_3, s_4)$, then $s_k = (s_2, s_3)$.)

## 5.2 Exploiting Passivity in the Transition Step

In order to perform selective updates over the belief factors $b_k$, we require a procedure which performs the transition step independently for each factor.[3] We obtain such a procedure by introducing two assumptions which allow us to modify the transition step (1) of the exact update rule. The assumptions guarantee that the transition step is performed exactly, in the sense of (1). However, as we will discuss shortly, the assumptions can be violated to obtain approximate belief states.

The first assumption, (A1), states that the clusters must be uncorrelated (i.e. there are no edges in $X^{t+1}$ between clusters), and the second assumption, (A2), states that the clusters must be disjoint. Formally, these are defined as follows:

(A1) $\quad \forall a : x_i^{t+1} \in C_k \rightarrow pa_a^{t+1}(x_i^{t+1}) \subseteq C_k$

(A2) $\quad \forall k \neq k' : C_k \cap C_{k'} = \emptyset$

Note that neither assumption implies the other. That is, it may be the case that (A1) is satisfied while (A2) is violated, and vice versa. Assuming both (A1) and (A2), we can reformulate (1) to

$$\hat{b}_k^{t+1}(s_k') \;=\; \eta_1 \sum_{\bar{s} \in S(pa_{a^t}(C_k))} T_k^{a^t}(\bar{s}, s_k') \prod_{k':[\exists x_i^{t+1} \in C_{k'} : x_i^t \in pa_{a^t}(C_k)]} b_{k'}^t(\bar{s}_{k'}) \qquad (6)$$

where $\eta_1$ is a normalisation constant and

$$T_k^a(\bar{s}, s_k') \;=\; \prod_{x_i^{t+1} \in C_k} P_a\big(x_i^{t+1} = (s_k')_i \mid pa_a(x_i^{t+1}) \leftarrow (\bar{s}, s_k')\big).$$

---

3. This also has the advantage that the belief factors can be updated in parallel, which is a useful feature considering that many platforms use parallel processing techniques.

This procedure performs the transition step independently for each belief factor $b_k$, hence they can be updated in any order and in parallel.

Assumption (A1) is what allows us to bring (1) into a form which updates the belief factors $b_k$ independently of each other. Specifically, (A1) allows us to define the cluster-based transition function $T_k^a$, which in turn enables the summation in (6). Assumption (A2), on the other hand, guarantees that the product in (6) is correct. In particular, it may be the case that $|\bar{s}_{k'}| < |C_{k'}|$ (i.e. there are fewer elements in $\bar{s}_{k'}$ than in $C_{k'}$) if there are variables in $C_{k'}$ which are not in $pa_{a^t}^t(C_k)$ (i.e. $x_i^{t+1} \in C_{k'}$ but $x_i^t \notin pa_{a^t}^t(C_k)$). In such cases, $b_{k'}^t$ is taken to be the marginal distribution over variables $x_i^{t+1} \in C_{k'}$ with $x_i^t \in pa_{a^t}^t(C_k)$, where (A2) guarantees that the marginalisation introduces no errors.

As mentioned previously, each assumption may be violated to obtain approximate belief states. However, there is an important distinction between (A1) and (A2) in this regard: If (A2) is violated, then (6) is still well-defined in the sense that it can still be executed, except that the product in (6) may degrade the accuracy of the results. This is in contrast to (A1), which is a *structural* requirement of $T_k^a$ in the sense that $T_k^a$ is ill-defined without (A1). This is since, if (A1) is violated, the variables in $C_k$ may have parents in $X^{t+1}$ which are not in $C_k$, in which case $pa_a(x_i^{t+1}) \leftarrow (\bar{s}, s_k')$ would be ill-defined. Thus, if (A1) is violated, we have to *enforce* it by modifying the distributions $P_a$ of all $x_i^{t+1} \in C_k$ to marginalise out all variables in $pa_{a^t}^{t+1}(x_i^{t+1})$ which are not in $C_k$, for all clusters $C_k$. This means that each variable has a separate distribution for every cluster which contains the variable, thereby possibly introducing an approximation error.

Given the modified transition step (6), we can exploit passivity to perform selective updates over the belief factors $b_k$. Recall from Section 4.1 that a variable $x_i^{t+1}$ is passive in $\Delta^a$ if there exists a set $\Phi_{a,i}$ of variables such that $x_i^{t+1}$ may change its value only if any of the variables in $\Phi_{a,i}$ changed its value. This causal connection can be used to decide whether or not the values of the variables in a cluster $C_k$ may have changed, in which case the corresponding belief factor $b_k$ should be updated. Theorem 1 provides the formal foundation:

**Theorem 1.** If (A1) and (A2) hold, and if all $x_i^{t+1} \in C_k$ are passive in $\Delta^{a^t}$, then

$$\forall s \in S : \hat{b}_k^{t+1}(s_k) = b_k^t(s_k).$$

*Proof.* Proof in Appendix A. □

Theorem 1 states that if the clusters $C_1, ..., C_K$ are disjoint and uncorrelated, and if all variables in cluster $C_k$ are passive in $\Delta^{a^t}$, then the transition step for the corresponding belief factor $b_k^t \rightarrow \hat{b}_k^{t+1}$ can be omitted without loss of information.

How does Theorem 1 translate into situations in which (A1) or (A2), or both, are violated? The key assumption is again (A1), which states that the clusters must be uncorrelated. As discussed earlier, we can enforce this by modifying the variable distributions $P_a$ in each cluster. However, if a passive variable $x_i^{t+1} \in C_k$ is correlated with a (passive or active) variable $x_j^{t+1} \in C_{k'}$, where $x_j^{t+1} \in pa_a^{t+1}(x_i^{t+1})$, then marginalising out $x_j^{t+1}$ in the distribution $P_a$ of $x_i^{t+1}$ will typically cause $x_i^{t+1}$ to lose its passivity, in the sense that it would no longer satisfy the clauses in Definition 3. Consequently, we would always have to perform the transition step for $C_k$, even if the unmodified variables in $C_k$ are all passive. This is problematic not only because of the unnecessary computations, but also because the modified distributions will introduce an error every time the transition step is performed.

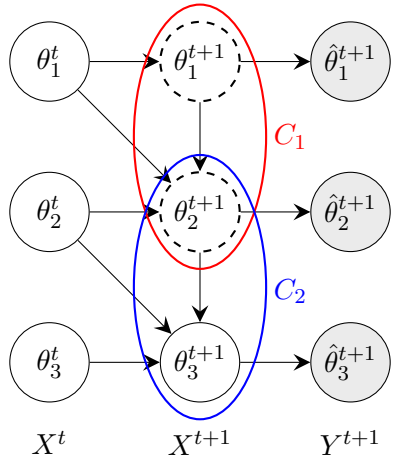

Figure 6: Robot arm DBN implementing the action $CW_3$. Dashed circles mark passive state variables. The coloured ellipses represent the clusters $C_1$ and $C_2$.

To alleviate this effect, one can check if there is a chance that the *unmodified* variables in the cluster would change their values. It can be shown that this is the case whenever there is a *causal path* from any active variable to a variable in the cluster:

**Definition 6** (Causal path). A *causal path* in $\Delta^a$, from an active variable $x_i^{t+1}$ to another variable $x_j^{t+1}$, is a sequence $\langle x^{(1)}, x^{(2)}, ..., x^{(Q)} \rangle$ such that $x^{(1)} = x_i^{t+1}, x^{(Q)} = x_j^{t+1}$, and for all for all $1 \leq q < Q$:

(i) $x^{(q)} \in X^{t+1}$

(ii) $(x^{(q)}, X^{(q+1)}) \in E_a$

(iii) $x^{(q+1)}$ is passive in $\Delta^a$ with respect to $x^{(q)}$

Intuitively, a causal path defines a *chain* of causal effects (such as between joints 1 and 3 in Example 1): since the active variable $x^{(1)}$ may have changed its value and $x^{(2)}$ is passive with respect to $x^{(1)}$, $x^{(2)}$ may also have changed its value; since $x^{(2)}$ may have changed its value and $x^{(3)}$ is passive with respect to $x^{(2)}$, $x^{(3)}$ may also have changed its value, etc. Hence, in the absence of observing these changes, the mere existence of a causal path from $x^{(1)}$ to $x^{(Q)}$ is reason to revise our beliefs about $x^{(Q)}$. Therefore, as a general update rule, we can omit the transition step $b_k^t \to \hat{b}_k^{t+1}$ if all unmodified variables in cluster $C_k$ are passive in $\Delta^{a^t}$, and if there is no causal path from any active variable in $\Delta^{a^t}$ to any variable in $C_k$. This is demonstrated in the following example:

**Example 5** (PSBF update rule in robot arm DBN). Let us again consider the robot arm from the previous examples. Figure 6 shows a DBN which implements the action $CW_3$. This action rotates joint 3 of the robot arm by $1°$ clock-wise (i.e. the joint orientation $\theta_3^{t+1}$ is a direct target of the action). Therefore, the variable $\theta_3^{t+1}$ is active while the variables $\theta_1^{t+1}$ and $\theta_2^{t+1}$ are passive (shown as dashed circles).

We use the clustering $C_1 = \{\theta_1^{t+1}, \theta_2^{t+1}\}$, $C_2 = \{\theta_2^{t+1}, \theta_3^{t+1}\}$ for reasons given in Example 4. Since $\theta_1^{t+1}$ is a parent of $\theta_2^{t+1}$, PSBF will have to enforce assumption (A1) by

---

**Algorithm 2** SKIPPABLECLUSTERS($\mathbb{C}, \Delta^a$)

---

1: **Input:** clustering $\mathbb{C} = \{C_1, ..., C_K\}$, DBN $\Delta^a$

2: **Output:** set of clusters $\mathbb{C}^* \subset \mathbb{C}$ which can be skipped in transition step

3: $\mathbb{C}^* \leftarrow \mathbb{C}$

4: $Q \leftarrow$ ORDEREDQUEUE($X^{t+1}$)

5: **while** $\mathbb{C}^* \neq \emptyset \wedge Q \neq \emptyset$ **do**

6:    $x_i^{t+1} \leftarrow$ NEXTELEMENT($Q$)

7:    $Q \leftarrow Q \setminus \{x_i^{t+1}\}$

8:    **if** $\neg$PASSIVE($x_i^{t+1}, \Delta^a$) **then**

9:       $\mathbb{C}^* \leftarrow \mathbb{C}^* \setminus \{C_k \in \mathbb{C}^* \mid x_i^{t+1} \in C_k\}$

10:       **for all** $x_j^{t+1} \in Q$ **do**

11:          **if** CAUSALPATH($x_i^{t+1}, x_j^{t+1}, \Delta^a$) **then**

12:             $\mathbb{C}^* \leftarrow \mathbb{C}^* \setminus \{C_k \in \mathbb{C}^* \mid x_j^{t+1} \in C_k\}$

13:             $Q \leftarrow Q \setminus \{x_j^{t+1}\}$

14: **return** $\mathbb{C}^*$

---

marginalising $\theta_1^{t+1}$ out of the variable distribution $P_a$ of $\theta_2^{t+1}$ in cluster $C_2$. While the modified variable distribution loses the passivity property (both clauses of Definition 3 are violated), the unmodified distribution of $\theta_1^{t+1}$ is still passive.

When performing the transition step, PSBF has to update the belief factor $b_2$ because the corresponding cluster $C_2$ contains the active variable $\theta_3^{t+1}$. However, since all variables in cluster $C_1$ are passive (there are no modified variables in $C_1$), and since there is no causal path from $\theta_3^{t+1}$ to any variable in $C_1$, PSBF can omit the update for the belief factor $b_1$. Intuitively, this makes sense since a change in the orientation of joint 3 cannot cause a change in the orientations of the preceding joints. Note that this corresponds to a saving of 50% in the transition step. ∎

Algorithm 2 defines a procedure which utilises this rule to find clusters for which the transition step can be skipped. The algorithm takes as inputs a clustering $\mathbb{C}$ and a DBN $\Delta^a$, and returns a set $\mathbb{C}^*$ of skippable clusters. It essentially searches through all active variables $x_i^{t+1}$ in $\Delta^a$ and removes all clusters $C_k$ from $\mathbb{C}$ which contain variables to which there is a causal path from $x_i^{t+1}$. The function ORDEREDQUEUE($X^{t+1}$) returns an ordered queue $Q$ with all variables in $X^{t+1}$. The performance of Algorithm 2 depends on the order of the queue. In our experiments, we obtained good performance by ordering the variables in descending order of their number of outgoing edges. The function NEXTELEMENT($Q$) returns the next element in the queue; the function PASSIVE($x_i^{t+1}, \Delta^a$) is defined in Algorithm 1; and the function CAUSALPATH($x_i^{t+1}, x_j^{t+1}, \Delta^a$) returns a logical true if and only if there is a causal path from $x_i^{t+1}$ to $x_j^{t+1}$ in $\Delta^a$.[4] Note that, given the invariance of passivity to process states (cf. Section 4.1), it suffices to call Algorithm 2 only once (in advance or as needed) to determine which of the clusters to omit in the transition step.

### 5.3 Efficient Incorporation of Observations

PSBF can perform the observation step similarly to the exact update rule (2), which conditions the propagated belief state $\hat{b}^{t+1}$ on the observation $o^{t+1}$ to obtain a fully updated belief state $b^{t+1}$. However, given the factored belief state representation used by PSBF, we require a procedure which respects this factorisation in the observation step. Assuming that (A1) and (A2) both hold, we can bring (2) into a form which updates the belief factors $b_k$ independently of each other

$$b_k^{t+1}(s_k') \;=\; \eta_2 \; \hat{b}_k^{t+1}(s_k') \sum_{\bar{s} \in S(pa_{a^t}^{t+1}(Y^{t+1}))\,:\,\bar{s}_k = s_k'} \Omega^{a^t}(\bar{s}, o^{t+1}) \prod_{k' \neq k\,:\,C_{k'} \cap pa_{a^t}^{t+1}(Y^{t+1}) \neq \emptyset} \hat{b}_{k'}^{t+1}(\bar{s}_{k'}) \tag{7}$$

where $\eta_2$ is a normalisation constant. Note that, analogously to (6), if there are variables in $C_{k'}$ which are not in $pa_{a^t}^{t+1}(Y^{t+1})$, then $\hat{b}_{k'}^{t+1}$ is taken to be the marginal distribution over $C_{k'} \cap pa_{a^t}^{t+1}(Y^{t+1})$. Assumption (A2) guarantees that the marginalisation introduces no errors. If (A1) and (A2) both hold, then the transition step (6) and observation step (7) produce *exact* belief states in the sense of (1) and (2), regardless of how many clusters were skipped in the transition step (cf. Theorem 1).

The observation step (7) updates *all* belief states and uses *all* observation variables in the process. In other words, it ignores the internal structure of the observation variables. However, it is clear that if the variables in a cluster $C_k$ are marginally independent of the observation variables $Y^{t+1}$ (this can be determined using *d-separation* (Geiger et al., 1989), or simply by checking if there is a directed path from $C_k$ to $Y^{t+1}$), then there is no need to perform the observation step for the corresponding belief factor $b_k$. This is expressed formally in Theorem 2:

**Theorem 2.** If all $x_i^{t+1} \in C_k$ are marginally independent of all $y_j^{t+1} \in Y^{t+1}$ in $\Delta^{a^t}$, then

$$\forall s \in S : b_k^{t+1}(s_k) = \hat{b}_k^{t+1}(s_k).$$

*Proof.* Proof in Appendix B. □

Theorem 2 states that if the variables in $C_k$ are independent of those in $Y^{t+1}$, then the observation step for $b_k$ can be skipped. However, even if $C_k$ is not independent of $Y^{t+1}$, it may be the case that the variables in $C_k$ depend only on a subset $Y_k \subset Y^{t+1}$ of the observation variables. Clearly, in such cases, it suffices to use $Y_k$ rather than $Y^{t+1}$ in the observation step. To account for this, we first note that the variables in $Y^{t+1}$ may be correlated with each other. To preserve the correlations, we subdivide $Y^{t+1}$ into clusters $\hat{C}_l \subseteq Y^{t+1}$ and introduce the following assumptions:

(A3)  $\forall a : y_j^{t+1} \in \hat{C}_l \rightarrow \left( pa_a(y_j^{t+1}) \cap Y^{t+1} \right) \subseteq \hat{C}_l$

(A4)  $\forall l \neq l' : \hat{C}_l \cap \hat{C}_{l'} = \emptyset$

---

4. A simple way to implement this function is to modify a standard graph search method (such as breath-first search) to check for (iii) in Definition 6, and to apply it to the variables in $X^{t+1}$ with edges $E_a$ from $\Delta^a$.

Assumptions (A3) and (A4) are analogous to (A1) and (A2), respectively, and essentially serve the same purposes for the observation step. To distinguish the clusters $C_k$ and $\hat{C}_l$, we sometimes refer to the former as *state cluster* and to the latter as *observation cluster*. Assuming that (A3) and (A4) both hold, we can redefine the observation step to

$$b_k^{t+1}(s_k') \;=\; \eta_2\, \hat{b}_k^{t+1}(s_k') \prod_{l:\,\hat{C}_l \cap Y_k \neq \emptyset} \; \sum_{\bar{s} \in S(pa_{a^t}^{t+1}(\hat{C}_l)):\, \bar{s}_k = s_k'} \Omega_l^{a^t}(\bar{s}, o^{t+1}) \prod_{k' \neq k\,:\, C_{k'} \cap pa_{a^t}^{t+1}(\hat{C}_l) \neq \emptyset} \hat{b}_{k'}^{t+1}(\bar{s}_{k'}) \qquad (8)$$

where

$$\Omega_l^a(\bar{s}, o^{t+1}) \;=\; \prod_{y_j^{t+1} \in \hat{C}_l} P_a\Big(y_j^{t+1} = (o_l^{t+1})_j \mid pa_a(y_j^{t+1}) \leftarrow (\bar{s}, o_l^{t+1})\Big)$$

and $Y_k \subset Y^{t+1}$ is the set of observation variables which are not marginally independent of the variables in $C_k$.

Given Theorem 2, one can see that (8) is equivalent to (7) if the observation variables are not clustered (or, equivalently, there is a single observation cluster $\hat{C}_l = Y^{t+1}$). However, it is important to note that if the observation variables are clustered (i.e. there are multiple observation clusters $\hat{C}_l$), then (8) is not necessarily equivalent to (7). To see this, it is helpful to compare the abstract formulations $\prod_{j=1}^m \sum_s \Omega_s(o_j)\, b_s$ and $\sum_s \prod_{j=1}^m \Omega_s(o_j)\, b_s$, where the former corresponds to (8) and the latter to (7). Therein, $(o_1, ..., o_m) \in O$ is an observation, $b_s$ is the probability of being in state $s \in S$, and $\Omega_s(o_j)$ is the probability of observing $y_j = o_j$ in $s$. These abstract formulations are equivalent for $m = 1$ or if $b_s = 1$ for some $s$, but in all other cases they may not be equivalent. Nonetheless, if we fix the number of observation variables $m$, then (8) approximates (7) closely as we increase the number of state variables $n$. Our experiments indicate that it often suffices to use just a few more state variables than observation variables in order to obtain good approximations.

Finally, to show that it suffices to perform the observation step for $b_k$ using only those clusters $\hat{C}_l$ whose variables are not independent of the variables in $C_k$, we observe that (8) is in fact a repeated application of (7) for every $\hat{C}_l$, where the updated belief factor $b_k^{t+1}$ is used in place of $\hat{b}_k^{t+1}$ in the subsequent application. Since every application has the same form as (7) (with $Y^{t+1} = \hat{C}_l$), we conclude that Theorem 2 holds, and hence the observation step can be skipped for clusters $\hat{C}_l$ which are independent of $C_k$.

## 5.4 Summary of PSBF

The preceding sections can be summarised as follows:

- **Representation**: The belief state $b^t$ is represented as a product of $K$ belief factors $b_k^t$, such that $b^t(s) = \prod_{k=1}^K b_k^t(s)$. Each belief factor $b_k^t$ is a probability distribution over the set $S(C_k)$, where $C_k \subseteq X^{t+1}$ is a cluster of correlated state variables.

- **Transition step**: The transition step $b_k^t \to \hat{b}_k^{t+1}$ is performed using (6), for all clusters $C_k$ which include active variables in $\Delta^{a^t}$, or to which there is a causal path from an active variable in $\Delta^{a^t}$. All other clusters are skipped.

- **Observation step**: The observation step $\hat{b}_k^{t+1} \to b_k^{t+1}$ is performed using (8), for all clusters $C_k$ which are dependent on the observation variables $Y^{t+1}$, using only those observation clusters $\hat{C}_l$ which are relevant for $C_k$. All other clusters are skipped.

---

**Algorithm 3** $\mathrm{PSBF}(a^t, o^{t+1}, (b_k^t)_{C_k \in \mathbb{C}} \mid \mathbb{C}, \hat{\mathbb{C}}, (\Delta^a)_{a \in A})$

---

1: **Input:** action $a^t$, observation $o^{t+1}$, belief factors $(b_k^t)_{C_k \in \mathbb{C}}$

2: **Parameters:** state clustering $\mathbb{C}$, observation clustering $\hat{\mathbb{C}}$, DBNs $(\Delta^a)_{a \in A}$

3: **Output:** updated belief factors $(b_k^{t+1})_{C_k \in \mathbb{C}}$

4: // *Transition step*

5: $\mathbb{C}^* \leftarrow \textsc{SkippableClusters}(\mathbb{C}, \Delta^{a^t})$

6: **for all** $C_k \in \mathbb{C}$ **do**

7:     **if** $C_k \in \mathbb{C}^*$ **then**

8:         $\hat{b}_k^{t+1} \leftarrow b_k^t$

9:     **else**

10:         **for all** $s_k' \in S(C_k)$ **do**

11:             $\hat{b}_k^{t+1}(s_k') \leftarrow \eta_1 \sum_{\bar{s} \in S(pa_{a^t}^t(C_k))} T_k^{a^t}(\bar{s}, s_k') \prod_{k': [\exists x_i^{t+1} \in C_{k'} : x_i^t \in pa_{a^t}^t(C_k)]} b_{k'}^t(\bar{s}_{k'})$

12: // *Observation step*

13: **for all** $C_k \in \mathbb{C}$ **do**

14:     $Y_k \leftarrow \left\{ y_j^{t+1} \in Y^{t+1} \mid \text{there is a directed path from } C_k \text{ to } y_j^{t+1} \text{ in } \Delta^{a^t} \right\}$

15:     **if** $Y_k = \emptyset$ **then**

16:         $b_k^{t+1} \leftarrow \hat{b}_k^{t+1}$

17:     **else**

18:         **for all** $s_k' \in S(C_k)$ **do**

19:             $b_k^{t+1}(s_k') \leftarrow \eta_2 \, \hat{b}_k^{t+1}(s_k') \prod_{\hat{C}_l \in \hat{\mathbb{C}} \, : \, \hat{C}_l \cap Y_k \neq \emptyset} \sum_{\bar{s} \in S(pa_{a^t}^{t+1}(\hat{C}_l)) \, : \, \bar{s}_k = s_k'} \Omega_l^{a^t}(\bar{s}, o^{t+1}) \prod_{k' \neq k \, : \, C_{k'} \cap pa_{a^t}^{t+1}(\hat{C}_l) \neq \emptyset} \hat{b}_{k'}^{t+1}(\bar{s}_{k'})$

20: **return** $(b_k^{t+1})_{C_k \in \mathbb{C}}$

---

Algorithm 3 provides a procedural specification of PSBF. The algorithm takes as inputs the action at time $t$, $a^t$, the subsequent observation at time $t+1$, $o^{t+1}$, and the belief factors at time $t$, $b_k^t$. The internal parameters are the state clustering $\mathbb{C}$, the observation clustering $\hat{\mathbb{C}}$, and the set of DBNs $(\Delta^a)_{a \in A}$ which define the process. Lines 4 to 11 implement the transition step while lines 12 to 19 implement the observation step. Note that it suffices to execute lines 5 and 14 once in advance (or on demand) and to remember the results for future reference. The algorithm returns the updated belief factors $b_k^{t+1}$.

## 5.5 Space and Time Complexity

A belief factor $b_k$ has one element $b_k(s_k)$ for each $s_k \in S(C_k)$.[5] Thus, the total space required to maintain $K$ belief factors $b_k$ is $\sum_{k=1}^{K} |S(C_k)|$. Furthermore, the size of the set $S(C_k)$ grows exponentially with the number of variables in $C_k$, hence the dominant growth factor in the space requirement is given by the largest cluster $C_k$ such that $|C_k| = \max_{k'} |C_{k'}|$. Therefore, the space complexity of PSBF is in $O(\exp \max_k |C_k|)$, hence the representation is feasible for reasonably small clusters $C_k$.

Similarly, the number of operations required to perform the transition and observation steps is in the order of $2 \sum_{k=1}^{K} |S(C_k)|$ in the worst case (i.e. all clusters need to be updated in both steps). Specifically, line 11 and line 19 in Algorithm 3 are each executed once for every $s_k \in C_k$. The dominant growth factor is again given by the largest cluster $C_k$, hence the time complexity of PSBF is in $O(2 \exp \max_k |C_k|) = O(\exp \max_k |C_k|)$. Note that this assumes that the analysis performed by lines 5 and 14 in Algorithm 3 is done in advance.

The above time complexity is for the *worst case*, in which all clusters need to be updated in the transition and observation steps. It is difficult to derive the time complexity for the average case because it is unclear what the average case is in terms of passivity. Even if we stipulate a certain average degree of passivity (e.g. 50% of all variables are passive), it would still be difficult to make a general statement about time requirements since this depends crucially on how the passive variables are distributed across the clusters. For example, even if a process has on average 90% passivity, if there is one active variable in each cluster then every cluster would need to be updated in the transition step. Thus, the only general statement we can make with regards to passivity is that the time complexity of PSBF can be refined to $O(\exp \max_{C_k \in \mathbb{C}^T \cup \mathbb{C}^O} |C_k|)$, where $\mathbb{C}^T$ and $\mathbb{C}^O$ include only those clusters that need to be updated in the transition and observation step, respectively.

## 5.6 Error Bounds

There are five possible sources of approximation errors in PSBF:

- If the clusters are correlated (i.e. (A1) or (A3) are violated)

- If the clusters are overlapping (i.e. (A2) or (A4) are violated)

- Generally in (8) if multiple observation clusters $\hat{C}_l$ are used

In the first two cases, the approximation error depends on the amount of correlation and overlap. If there is only little correlation and overlap between the clusters, then the approximation error can be expected to be small. Conversely, if the clusters are strongly correlated and overlapping, then the approximation error can be expected to be large.

Boyen and Koller (1998) provide a useful analysis of the error bound of any filtering method which uses a factored belief state representation. Since PSBF uses a factored representation, their analysis applies directly to PSBF. The purpose of this section is to restate the main result of their analysis in the context of our work.

Their analysis uses the concept of *relative entropy* (Kullback & Leibler, 1951) as a measure of similarity for belief states:

---

5. In practice, it suffices to store only $|S(C_k)| - 1$ elements, but this is irrelevant in our analysis.

**Definition 7** (Relative entropy). Let $\phi$ and $\psi$ be two probability distributions defined over a set $X$. The *relative entropy* from $\phi$ to $\psi$ is defined as

$$KL(\phi||\psi) = \sum_{x \in X} \phi(x) \ln \frac{\phi(x)}{\psi(x)}$$

where $\phi(x) > 0 \Rightarrow \psi(x) > 0$.

Similar to Boyen and Koller (1998), we define the approximation error incurred by PSBF relative to the exact belief state. However, since we consider a decision process with multiple actions $a \in A$ (represented by the DBNs $\Delta^a$), we define the error for each action respectively:

**Definition 8** (Approximation error). Let $b$ be an exact belief state and $\tilde{b}$ be the approximation by PSBF. After taking action $a$, let $b'$ be the exact update of $b$ (using (1) and (2)) and $\tilde{b}'$ be the PSBF-update of $\tilde{b}$ (using (6) and (8)). Furthermore, let $\check{b}'$ be the exact update of $\tilde{b}$ (using (1) and (2)). We say that PSBF incurs error $\epsilon^a$ in $\Delta^a$ relative to $b'$ if

$$KL(b'||\tilde{b}') - KL(b'||\check{b}') \leq \epsilon^a.$$

The analysis also relies on the concept of *mixing rates*. Intuitively, the mixing rate $\gamma^a$ of a DBN $\Delta^a$ quantifies the degree of stochasticity in $\Delta^a$. It depends on the mixing rates $\gamma_k^a$ of the individual clusters $C_k$:

**Definition 9** (Mixing rate). The *mixing rate* of a cluster $C_k \subset X^{t+1}$ in $\Delta^a$ is defined as

$$\gamma_k^a = \min_{s', s'' \in S} \sum_{s \in S(C_k)} \min\left[ T_k^a(s', s), T_k^a(s'', s) \right].$$

If all $C_k$ satisfy (A1) and (A2), and if all observation variables $Y^{t+1}$ are in one observation cluster, then the mixing rate of $\Delta^a$ is given by $\gamma^a = (\min_k \gamma_k^a / r)^q$ where each cluster $C_k$ depends on at most $r$ and influences at most $q$ other clusters $C_{k' \neq k}$ (Boyen & Koller, 1998). In the worst case (that is, all (A1–A4) are violated), the minimal mixing rate is given by $\gamma_k^a$ for the single cluster $C_k = X^{t+1}$.

Finally, the main result in the work of Boyen and Koller (1998), here restated in the context of our work in Theorem 3, essentially states that the approximation error of PSBF (measured in terms of relative entropy) is bounded by the mixing rates of the process:

**Theorem 3** (Boyen & Koller, 1998). Let $b^t$ be an exact belief state and $\tilde{b}^t$ be the approximation by PSBF using clusters $C_k$. Then, for any $t$ with states $\vec{s} = (s^0, ..., s^t)$ and actions $\vec{a} = (a^0, ..., a^{t-1})$, we have

$$E_{o^1, ..., o^t}\left[ KL(b^t || \tilde{b}^t) \right] \leq \frac{\max_{a \in \vec{a}} \epsilon^a}{\min_{a \in \vec{a}} \gamma^a}$$

where the expectation $E$ is taken over all possible sequences of observations $o^1, ..., o^t$ with probabilities $P(o^1, ..., o^t) = \prod_{\tau=0}^{t-1} \Omega^{a^\tau}(s^{\tau+1}, o^{\tau+1})$, and where $\epsilon^a$ and $\gamma^a$ are defined as above.

| Process size | # of $x$ vars ($n$) | # of $y$ vars ($m$) | # of states ($|S|$) | # of obs. ($|O|$) |
|:---:|:---:|:---:|:---:|:---:|
| S | 10 | 3 | > one thousand | 8 |
| M | 20 | 6 | > one million | 64 |
| L | 30 | 9 | > one billion | 512 |
| XL | 40 | 12 | > one trillion | 4096 |

Table 1: Synthetic process sizes. All variables are binary.

## 6. Experimental Evaluation

We evaluated PSBF in two experimental domains: In Section 6.1, we evaluated PSBF in synthetic (i.e. randomly generated) processes with varying sizes and degrees of passivity. In Section 6.2, we evaluated PSBF in a simulation of a multi-robot warehouse system. A brief summary of the experimental results is given in Section 6.3.

### 6.1 Synthetic Processes

We first evaluated PSBF in a series of synthetic processes. PSBF is compared with a selection of alternative methods, including PF (Gordon et al., 1993), RBPF (Doucet et al., 2000), BK (Boyen & Koller, 1998), and FF (Murphy & Weiss, 2001); see Section 2 for a discussion of these methods. The algorithms were implemented in Matlab 7.13, where we used the Matlab toolbox BNT (Murphy, 2001) to implement BK and FF.

#### 6.1.1 Specification of Synthetic Processes

We generated synthetic processes of four different sizes which are specified in Table 1. Each process was generated as follows:

First, each variable $x_i^{t+1}$ is chosen to be passive with probability $p$, in which case we also add the edge $(x_i^t, x_i^{t+1})$. We refer to $p$ as the *degree of passivity*. To sample further edges from $X^t/X^{t+1}$ to $X^{t+1}$, we generate a mixture of Gaussians $\mathbb{G}$ using Algorithm 4 (see Appendix C). Figure 7 shows an example of $\mathbb{G}$ generated for a process of size M. The set $\mathbb{G}$ is used to produce "areas" of correlated variables (i.e. the Gaussians), which will then constitute natural candidates for state clusters.

Let $\omega$ be the vector of maximum densities for each Gaussian in $\mathbb{G}$, and let $\delta_i$ be the vector of densities at value $i \in \mathbb{N}$. Then, for every combination of $i$ and $j$, the edge $(x_i^t, x_j^{t+1})$ is added with probability equal to the maximum element in $\delta_i \delta_j / \omega^2$, in which all operators are point-wise. If $x_i^{t+1}$ was chosen to be passive, then the edge $(x_i^t, x_j^{t+1})$ is only added if $i < j$. In that case, we also add the edge $(x_i^{t+1}, x_j^{t+1})$. Edges $(x_i^{t+1}, x_j^{t+1})$ are added similarly for each $i < j$,[6] where we also add the edge $(x_i^t, x_j^{t+1})$ for passive $x_j^{t+1}$. To ensure that every variable has an effect in the generated process, each $x_i^t$ is connected to at least one $x_j^{t+1}$ (adding $(x_i^t, x_i^{t+1})$ if necessary) and each $x_j^{t+1}$ has at least one parent in $X^t$ or $X^{t+1}$ (adding $(x_j^t, x_j^{t+1})$ if necessary). Finally, edges $(x_i^{t+1}, y_j^{t+1})$ are added with probability 0.1, for each $i, j$, while ensuring that each $y_j^{t+1}$ has at least one parent in $X^{t+1}$.

---

6. The condition $i < j$ in both cases is to ensure that the resulting DBN is acyclic.

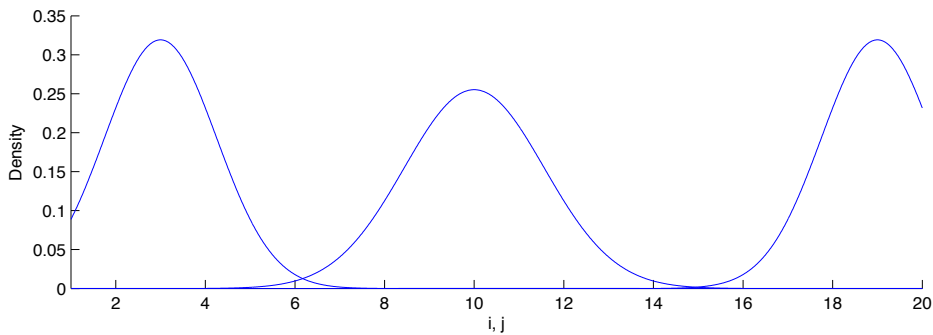

Figure 7: Example of mixture of Gaussians generated for a process of size M and consisting of three Gaussians. The closer two variables $x_i^{t/t+1}$ and $x_j^{t+1}$ are under the peak of a common Gaussian, the higher the probability that an edge will be added between them.

All variables in the process are binary. Passive variables are assumed to be passive with respect to all of their parents in $X^t$. The distributions $P_a$ of $x_i^{t+1} \in X^{t+1}$ are generated uniformly randomly without bias. For passive variables $x_i^{t+1}$, we modify $P_a$ to satisfy clause (ii) in Definition 3. The distributions $P_a$ of $y_j^{t+1} \in Y^{t+1}$ are generated with each probability sampled uniformly from either $[0.0, 0.2]$ or $[0.8, 1.0]$, to obtain meaningful observations.

Finally, every process consists of two actions. These are obtained by randomly choosing between one and three variables $x_i^{t+1}$ whose distributions $P_a$ are resampled as above and edges from $X^t$ added with probability 0.1 (passive variables chosen in this way are no longer passive). During simulations, these actions are chosen uniformly randomly.

Each process starts in a random initial state, and all algorithms are tested on the same sequence of processes, initial states, chosen actions, and random numbers.

### 6.1.2 Clustering Methods

We used three different clustering methods, denoted ⟨pc⟩, ⟨moral⟩, and ⟨modis⟩. The methods were applied to the variables in $X^{t+1}$ without edges involving $X^t$ or $Y^{t+1}$:

- ⟨pc⟩ drops the directions of the edges (i.e. for any edge $x_i^{t+1} \rightarrow x_j^{t+1}$ it ads the reverse edge $x_j^{t+1} \rightarrow x_i^{t+1}$) and puts all variables between which there is a (undirected) path into one cluster. By definition, the resulting clusters satisfy all assumptions (A1–A4).

- ⟨moral⟩ connects all parents of a variable and drops the directions (it "moralises" the variables) and then extracts clusters of fully connected variables ("maximum cliques"). The resulting clusters may not satisfy any of the assumptions (A1–A4).

- ⟨modis⟩ is similar to ⟨moral⟩ but truncates the resulting clusters to make them disjoint (clusters are removed if they become a subset of another cluster). By definition, the resulting clusters satisfy (A2/A4), but not necessarily (A1/A3).

As an example, consider Figure 5 from Section 5.1. Here, ⟨pc⟩ would produce the cluster $C_1$ from Figure 5b, since all variables are connected by an undirected path. Furthermore, $\langle moral \rangle$ would produce the two clusters $C_1$ and $C_2$ from Figure 5c, which correspond to the two maximum cliques after moralising the variables in $X^{t+1}$. Finally, $\langle modis \rangle$ would produce the cluster $C_1$ from Figure 5c and the cluster $C_3$ from Figure 5a.

PSBF used the same clustering method to generate clusters of state variables ($C_k$) and observation variables ($\hat{C}_l$). Moreover, PSBF enforced (A1/A3) whenever necessary by modifying the variable distributions as described in Section 5.1.

### 6.1.3 Accuracy

In order to compare the accuracy of the tested algorithms, we computed the relative entropy (cf. Definition 7) from exact belief states obtained using the exact update rule (cf. Definition 1) to the approximate belief states produced by the tested algorithms. However, since exact belief states and relative entropy are hard to compute for large processes, we were able to compare the accuracy of algorithms in processes of size S only. All algorithms were initialised with uniform belief states, or uniformly sampled particles.

We first compared the accuracy of PSBF and BK, since they use the same factorisation in their belief state representations. Figure 8 shows the relative entropy of PSBF and BK averaged over 1000 processes with 0%, 20%, 40%, 60%, 80%, and 100% passivity, respectively. The results show that PSBF $\langle pc/modis \rangle$ produced a lower relative entropy (i.e. higher accuracy) than BK $\langle pc/modis \rangle$, and that PSBF $\langle moral \rangle$ produced a relative entropy comparable to that of BK $\langle moral \rangle$. This indicates that violations of (A2/A4) introduce smaller errors than violations of (A1/A3). Note that PSBF and BK had the same convergent behaviour in their relative entropy, which shows that the approximation error due to the factorisation was bounded, as discussed in Section 5.6. This is interesting since PSBF and BK obtain approximation errors from the factorisation in different ways: PSBF loses accuracy by modifying the variable distributions to ensure that the state clusters are independent (cf. Section 5.2), while BK loses accuracy by marginalising out the original factorisation after the inference (i.e. the "projection step"; cf. Section 2.1). Nevertheless, as shown in our results, the resulting approximation errors were bounded in both cases, with similar convergence.

Note that the relative entropy of both methods increased with the degree of passivity in the process. This is explained by the fact that higher passivity implies higher determinacy and, therefore, lower mixing rates (cf. Definition 9), which are a crucial factor in the error bounds of PSBF and BK (cf. Theorem 3). Finally, note that PSBF did not produce exact belief states (i.e. zero relative entropy) when using $\langle pc \rangle$ clustering, despite the fact that the clusters generated by $\langle pc \rangle$ satisfy all assumptions (A1–A4). However, as discussed in detail in Sections 5.3 and 5.6, another possible source of approximation errors is if multiple observation clusters are used, which was often the case when using $\langle pc \rangle$ to produce observation clusters.

To compare the accuracy of PF/RBPF with PSBF/BK, the number of samples used in PF/RBPF was chosen automatically in each process such that they required approximately as much time per belief update as PSBF $\langle moral \rangle$ and BK $\langle moral \rangle$, respectively. In our experiments, this meant that PF (RBPF) was only able to process between 100 and 300 (20 and 50) samples. However, since each process has over 1000 states, this was not nearly enough to represent a uniform belief state. Hence, PF/RBPF produced much higher relative entropy than PSBF/BK. Moreover, the fact that the processes have very high variance means that PF/RBPF would require many more samples to achieve the same accuracy as PSBF/BK (as shown in the next section). One would expect that this latter issue was alleviated by the use of exact inference in RBPF (cf. Section 2.1). However, this is only the case if much of the variance in the process can be captured in the marginal distributions used in the particles in RBPF. In contrast, our synthetic processes exhibit high variance across all variables, and our automatic grouping[7] of state variables into "sampled" and "exact" variables still contained much variance in the sampled variables. Hence, RBPF required significantly more samples than the number it could process in the time provided.

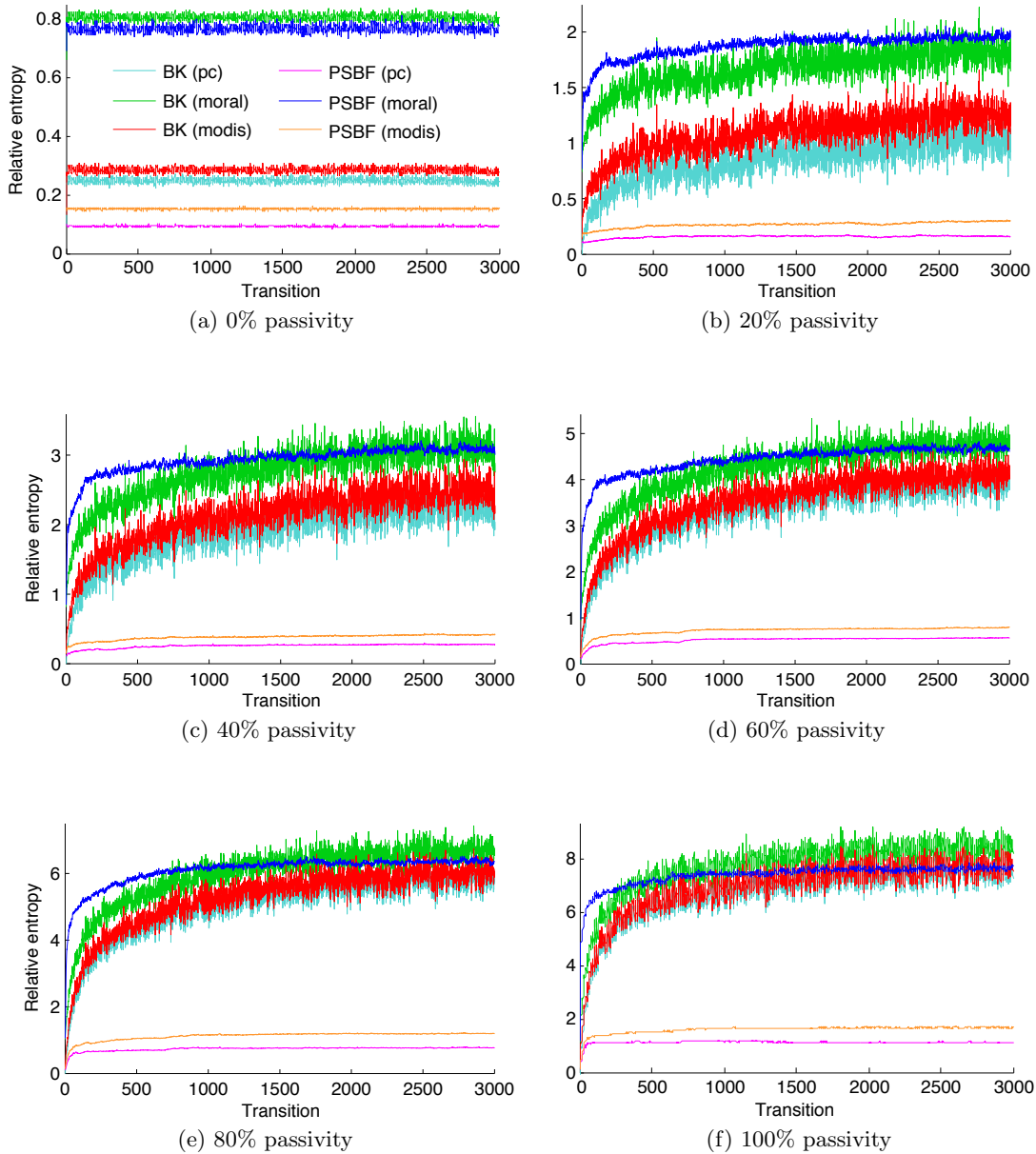

Figure 8: Accuracy results for PSBF and BK. Plots show relative entropy from exact to algorithms' belief states (lower is better). Results are averaged over 1000 processes of size S ($n = 10, m = 3$), where on average 0%–100% of non-target variables were passive (cf. Section 6.1.1). PSBF/BK used clustering methods $\langle pc \rangle$, $\langle moral \rangle$, and $\langle modis \rangle$.

Finally, in order to compare the accuracy of FF with PSBF/BK, the number of iterations used in FF (more precisely, the number of iterations in loopy belief propagation; cf. Murphy & Weiss, 2001) was chosen automatically in each process such that FF required approximately as much time per belief update as PSBF $\langle moral \rangle$ and BK $\langle moral \rangle$, respectively. However, while FF was often able to perform several iterations in the provided time, the resulting relative entropy was again substantially higher than that of PSBF/BK. The problem is that FF was designed for a specific class of DBN topologies, namely those containing no edges within $X^{t+1}$ (called "regular" DBNs by Murphy & Weiss, 2001). This is what allows FF to use a fully factored representation of belief states, in which each variable is its own belief factor. However, the processes used in our experiments have high intra-correlation between state variables (i.e. many edges in $X^{t+1}$), especially with increasing passivity. These correlations cannot be captured in the belief state representation of FF, resulting in a significantly higher relative entropy than PSBF/BK.

### 6.1.4 Timing

We measured computation times in processes of sizes S, M, L, XL with passivities of 25%, 50%, 75%, 100%, respectively. PSBF and BK used $\langle moral \rangle$ clustering, which seemed most appropriate for a fair comparison since it produced consistently similar accuracy for both algorithms. The number of samples used in PF was chosen automatically in each process such that PF achieved an average accuracy approximately as good as that of PSBF and BK, respectively, in the final 20% of the process. As this involved computing exact belief states and relative entropies, we were able to use PF in processes of size S only. We omit RBPF and FF in this section as they were shown in the previous section to be unsuitable for the processes we consider. PSBF was tested with 1, 2, and 4 parallel processes, which were allocated approximately the same number of belief factors.

Figures 9a – 9d show the times for 1000 transitions averaged over 1000 processes, and Figure 9e shows the average percentage of belief factors that were updated in the transition and observation steps of PSBF. The timing reported for PSBF includes the time taken to modify variable distributions (in case of overlapping clusters) and to detect skippable clusters in the transition and observation steps, both of which were done once in advance for each action. The results show that PSBF was able to minimise the time requirements significantly by exploiting passivity. First, we note that there were only marginal gains from 25% to 50% passivity, despite the fact that PSBF updated 14% fewer clusters in the transition step. This is because these clusters were mostly very small. However, there were significant gains from 50% to 75% passivity with average speed-ups of 11% (S), 14% (M), 15% (L), 18% (XL), and

---

7. It is an open question how to group state variables into "sampled" and "exact" variables (Doucet et al., 2000). We used a simple heuristic whereby the set of sampled variables contained all variables $x_i^{t+1}$ that had no parents in $X^{t/t+1}$ or none other than $x_i^t$. The remaining variables in $X^{t+1}$ constituted the set of exact variables. To ensure that the resulting grouping was valid for all actions (i.e. DBNs) in a process, we considered edges in all involved DBNs; that is, we performed the grouping over the union of $E_a$ for all $a$. Moreover, to improve efficiency, we further subdivided the set of exact variables into clusters of variables that were connected by undirected edges in $X^{t+1}$ without edges involving the sampled variables.

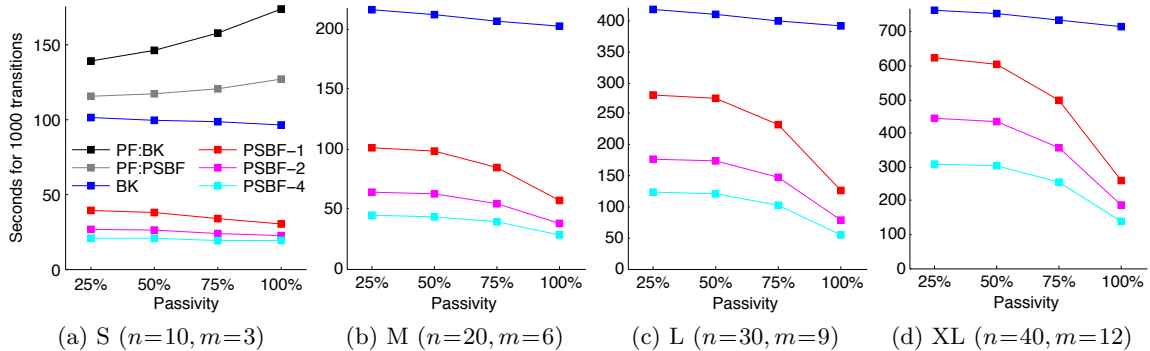

(a) S ($n=10, m=3$)    (b) M ($n=20, m=6$)    (c) L ($n=30, m=9$)    (d) XL ($n=40, m=12$)

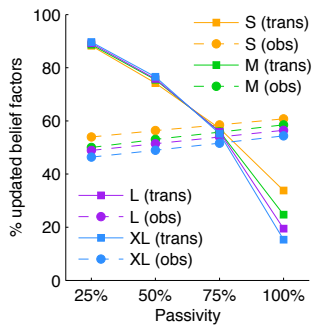

(e) Updated belief factors

Figure 9: Timing results. (a–d) Average number of seconds required for 1000 transitions on a UNIX dual-core machine with 2.4 GHz, for sizes S, M, L, XL. Passivity of $p\%$ means that on average $p\%$ of non-target variables were passive (cf. Section 6.1.1). PSBF and BK used ⟨moral⟩ clustering. PF was optimised for binary variables and used number of samples to achieve accuracy of PSBF and BK, respectively. PSBF was run with 1 (PSBF-1), 2 (PSBF-2), 4 (PSBF-4) parallel processes. (e) Average percentage of belief factors which were updated in the transition and observation steps, respectively.

from 75% to 100% passivity with further average speed-ups of 11% (S), 33% (M), 46% (L), 49% (XL). This shows that the computational gains can grow significantly with both the degree of passivity and the size of the process.

Our results show that PSBF consistently outperformed BK in all process sizes. There are two main computational savings in PSBF relative to BK: firstly, by skipping over belief factors in the transition and observation steps, and secondly, by not having to perform a potentially expensive projection step to restore the original factorisation after the inference. However, while the times of both algorithms grew exponentially in the size of the process, we note that the relative difference between PSBF and BK decreased significantly for lower degrees of passivity. This is an instance of "No Free Lunch" (see Section 7 for a discussion), which means that PSBF performs best in processes with high passivity but can suffer in performance in processes that lack passivity. Specifically, the computational overhead of modifying variable distributions and detecting skippable belief factors does not amortise as effectively in large processes with low passivity. Furthermore, with low passivity, PSBF often has to perform full transition and observation steps (i.e. update all belief factors in each step), which can be costly in large processes.

How were BK and PF affected by passivity? Not surprisingly, the performance of BK was nearly unaffected by the increasing degrees of passivity. The junction tree algorithm used in BK benefited marginally from an increased sparsity in the process, but the computational gains were minimal. We were at first unable to use PF as it required too many samples (between 10k and 200k) to achieve comparable accuracy to PSBF/BK, due to the very high variance in the processes. In order to investigate the effect of passivity on PF, we implemented a version of PF which was strictly optimised for binary variables. Interestingly, we found that passivity had an adverse effect on the performance of PF, requiring it to use exponentially more samples with increased passivity (see Figure 9a). This makes sense if we view PF as a factored approximation method (such as PSBF and BK) which means that the analysis in Section 5.6 applies. However, because PF puts all variables into a single cluster (since it is not *actually* a factored method), the mixing rate of the process will be much lower than for PSBF and BK (as discussed in Section 5.6) and, thus, the error bounds are less tight. To compensate for this, PF requires significantly more samples for increased passivity.

## 6.2 Multi-robot Warehouse System

In this section, we demonstrate how passivity can occur naturally in a more complex system and how PSBF can exploit this to accelerate the filtering task. To this end, we consider a multi-robot warehouse system in the style of Kiva (Wurman et al., 2008), in which the robots' task is to transport goods within the warehouse (cf. Figure 10a).

### 6.2.1 Specification of Warehouse System

Figure 10b shows the initial state of the warehouse simulation. The warehouse consists of 2 workstations (W1, W2), 4 robots (R1–R4), and 16 inventory pods (I1–I16). Each robot can move forward and backward, turn left and right, load and unload an inventory pod (if positioned under the pod), or do nothing. As in Kiva, robots can move under inventory pods unless they are carrying a pod, in which case the other pods become obstacles. The move and turn operations are stochastic in that the robot may move/turn too far (3% chance) or do nothing (2% chance). Each robot possesses two sensors, one telling it which inventory pod it has loaded (if any) and one for the direction it is facing. The direction sensor is noisy in that a random direction may be reported (3% chance).

Each robot maintains a list of tasks in the form of "Bring inventory pod I to workstation W" (yellow area around W) and "Bring inventory pod I to position (x,y)". How these tasks are executed depends on the control mode, of which we use two in our simulations:[8]

---

8. Our control modes are ad hoc and often make suboptimal decisions. However, we found that current solution techniques for (DEC-)POMDPs, including approximate methods, were infeasible in this setting. Nonetheless, the quality of the decisions made by our control modes largely depends on the accuracy of the belief states, hence it is important that the belief states are updated accurately. Therefore, the control modes were sufficient for our purposes.

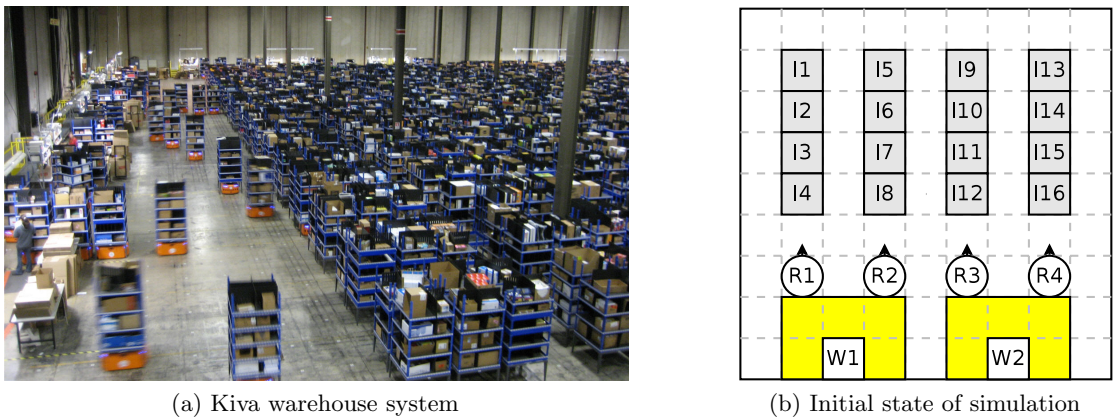

(a) Kiva warehouse system      (b) Initial state of simulation

Figure 10: (a) Kiva warehouse system (image reproduced from D'Andrea & Wurman, 2008). Robots (orange coloured) transport shelfs with goods to and from workstations. (b) Initial state of the warehouse simulation. The warehouse consists of 2 workstations (W1, W2), 4 robots (R1–R4), and 16 inventory pods (I1–I16).

**Centralised mode:** A central controller maintains a belief state $b^t$ about the state of the warehouse system. At each time $t$, it samples 100 states from $b^t$ and removes all duplicate states, resulting in the set $\hat{S} = \{\hat{s}_1, \hat{s}_2, ...\}$. It then resamples a state $\hat{s}^* \in \hat{S}$ with probabilities $w(\hat{s}^*) = b^t(\hat{s}^*)/\sum_q b^t(\hat{s}_q)$. Based on $\hat{s}^*$ and the current task of each robot, it performs an $A^*$ search (Hart, Nilsson, & Raphael, 1968) (with Manhattan distance) in the space of joint actions to find the optimal action for each robot. After executing their actions, the robots send their sensor readings to the controller, and the controller updates its belief state using the sensor readings.

**Decentralised mode:** Each robot maintains its own belief state and there is no communication between the robots. The only knowledge the robots have about each other are their current tasks, communicated by the task allocation module. At each time $t$, each robot samples the set $\hat{S}$ and state $\hat{s}^*$ as is done in the centralised mode. Treating the other robots as static obstacles, it performs an $A^*$ search based on $\hat{s}^*$ and its current task to find an action $a^t$. This is repeated for each other robot $r$ in all states $\hat{s}_q \in \hat{S}$, resulting in actions $a_{r,q}$ which are used to obtain distributions $\pi_r : A \to [0,1]$ ($A$ is the set of all actions) with $\pi_r(a) = \sum_{q\,:\,a_{r,q}=a} w(\hat{s}_q)$. The robot then executes its action $a^t$ and updates its belief state using its sensor readings and the distributions $\pi_r$ to average over the other robots' actions.

The tasks are generated by an external scheduler in time intervals sampled from $U[1, 10]$. Each generated task is assigned to one of the robots through a sequential auction (Dias, Zlot, Kalra, & Stentz, 2006). The robots' bids are calculated as their total number of steps needed to solve all of their current tasks and the auctioned task (in a simplified model in which the other robots are removed), averaged over all states in $\hat{S}$. The robot with the lowest bid is assigned the task.

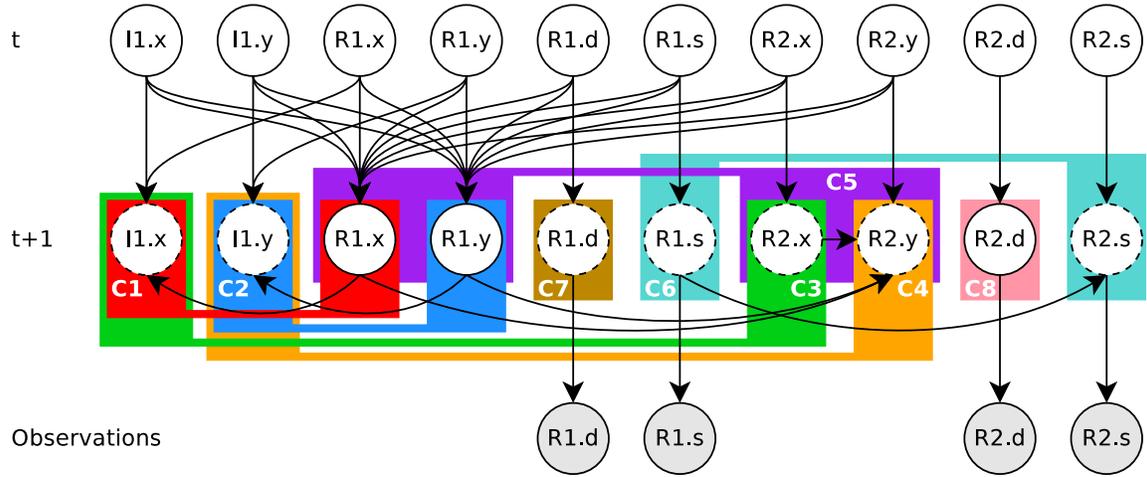

Figure 11: Example DBN of a smaller warehouse system consisting of only one inventory pod (I1) and two robots (R1, R2). The DBN implements the joint action in which R1 moves and R2 turns. Dashed circles mark passive state variables. The coloured areas represent the state clusters $C_1$ to $C_8$.

### 6.2.2 DBN Topology and Clustering

Figure 11 shows an example DBN for a smaller warehouse with one inventory pod and two robots. Each inventory pod I is represented by two variables, I.x and I.y, which correspond to the x and y position of the inventory pod. Each robot R is represented by four variables: R.x/R.y for its x/y position, R.d for its direction, and R.s for its status. The status of a robot R is either R.s=0 (unloaded) or R.s=I (loaded with inventory pod I). Constants such as the size of the warehouse and the positions of the workstations are omitted in the DBN.

There are four types of clusters: The I-clusters (C1–C4) preserve the correlation that if R is loaded with I, then I must always have the same position as R (there are two I-clusters for each (I,R) pair); The R-clusters (C5) and S-clusters (C6), respectively, preserve the correlation that no two robots can have the same position or carry the same inventory pod (there is one R/S-cluster for each (R$a$,R$b$) pair with $a > b$); And, finally, the D-clusters (C7, C8). PSBF uses singleton observation clusters (i.e. one cluster for each observation variable).

There are some differences between the DBNs for the centralised and decentralised modes (Figure 11 uses the centralised mode). In the centralised mode, there is one DBN for each action combination of the robots. Since the controller observes all R.s noise-free, it can add edges from R.x/R.y to I.x/I.y if R.s=I or remove them otherwise to simplify the inference (thus, in Figure 11, R1 is loaded with I1 and R2 is unloaded). In the decentralised mode, each robot only observes its own sensor readings, hence it can add or remove edges only for itself, while edges for all other robots must be permanently added. This also means that the other robots' status variables (R.s) must be linked to all I.x/I.y and, therefore, included in the I-clusters (to preserve the correlation that I must have the same position as R if R is loaded with I). Moreover, since each robot only knows its own action, there is one DBN for each of its own actions, and all variables associated with the other robots are active (the distributions $\pi_r$ defined in the previous section are used to average over their actions).

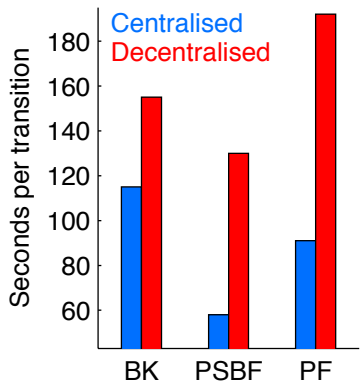

Figure 12: Results of the warehouse simulation, using the centralised and decentralised control modes. Timing measured on a UNIX dual-core machine with 2.4 GHz and averaged over 20 different simulations with 100 transitions each.

### 6.2.3 RESULTS

We implemented PSBF, BK, and PF in C#, using the framework Infer.NET (Minka, Winn, Guiver, & Knowles, 2012) to implement BK. This allowed BK to exploit sparsity in the process and offered improved memory handling. PSBF was optimised for sparsity in (6) and (8), respectively, by summing over states $\bar{s}$ for which all $b^t_{k'}/\hat{b}^{t+1}_{k'}$ are positive. PF naturally benefits from sparsity as it allows it to concentrate the samples on fewer states. The number of samples used in PF was set in such a way that the controller decisions were invariant of the random numbers used in the sampling process of PF. This was done to ensure that the results were repeatable. Finally, to maintain sparsity in the process, each probability in the belief states lower than 0.01 was set to 0. All tested algorithms were initialised with an exact belief state, shown in Figure 10b.

Figure 12 shows the time per transition averaged over 20 different simulations with 100 transitions each. The timing reported for PSBF includes the time needed to modify variable distributions (for overlapping clusters) and to detect skippable belief factors for the transition and observation steps, both of which were done once on demand for every previously unseen DBN. In the centralised mode, PSBF was able to outperform BK on average by 49% and PF by 36%. PF needed 20,000 samples to produce consistent (i.e. repeatable) results. In the decentralised mode, PSBF outperformed BK on average by 17% and PF by 32%. PF now needed 45,000 samples to produce consistent results, due to the increased variance in the process. All differences were statistically significant, based on paired t-tests with a 5% significance level. Note that PSBF and BK were slower in the decentralised mode since the corresponding DBNs had much higher inter-connectivity. In addition, PSBF updated more belief factors since there were more active variables.

As expected, PSBF was able to exploit the high degree of passivity in the process to accelerate the filtering task. In many cases, this meant that PSBF needed to update less than half of the belief factors. Precisely how many belief factors had to be updated depends on the performed action. To illustrate this, consider the smaller warehouse DBN shown in Figure 11 (for the centralised mode), in which R1 is moving and R2 is turning. Here, R1.x, R1.y, and R2.d are active variables while all other variables are passive (dashed circles), corresponding to a passivity of 70%. In this DBN, PSBF updates the belief factors corresponding to clusters C1, C2, C5, and C8, since they each contain active variables, and it also updates the belief factors for C3 and C4, since there are directed paths from active variables (R1.x and R1.y) to each of them. Therefore, the only factors which are not updated are for C6 and C7. Now consider the full warehouse in our experiment, which contains 16 inventory pods and 4 robots, resulting in 48 variables with 128 I-clusters, 6 R-clusters, 6 S-clusters, and 4 D-clusters. Assume a similar situation in which one robot moves with an inventory pod, say R4 with I1, while the R1–3 turn. In this case, PSBF updates only 3 of 6 R-clusters (those containing R4), 0 of 6 S-clusters (since no status change), 3 of 4 D-clusters (for R1–3), and 38 of 128 I-clusters (32 I-clusters containing R4 plus 6 I-clusters from R1–3 for I1), amounting to a total saving of 69.44% of belief factors which do not need to be updated.

The number of states in the warehouse system (including invalid states) exceeded $10^{45}$ states. Therefore, we were unable to compare the accuracy of the tested algorithms in terms of relative entropy. Instead, we compared their accuracy based on the results of the task auctions and the number of completed tasks by the end of each simulation. This gives a good indication of the algorithms' accuracy, since both the outcome of the auction and the number of completed tasks depend on the accuracy of the belief states. In the centralised mode, the algorithms generated over 95% identical task auctions and completed 15.7 (BK), 15.5 (PSBF), and 15.2 (PF) tasks on average. In the decentralised mode, they generated over 93% identical auctions and completed 12.1 (BK), 12.2 (PSBF), and 11.7 (PF) tasks on average. In both modes, none of these differences were statistically significant. Therefore, this indicates that PSBF achieved an accuracy similar to that of BK and PF.

## 6.3 Summary of Experimental Evaluation

The experimental results show that PSBF produces belief states with competitive accuracy: In the synthetic processes, PSBF achieved an accuracy which on average was better or comparable to the accuracy of the alternative methods. In the warehouse system, PSBF was able to complete a statistically equivalent number of tasks as compared to the other methods, which indicates that its accuracy was equivalent or comparable.

Furthermore, the experimental results show that PSBF performed the belief updates significantly faster than the alternative methods: In the synthetic processes, PSBF using no parallel processes outperformed BK by up to 64% in the largest process (XL), while PF took too much time to achieve an accuracy comparable to PSBF. In particular, the results show that the computational gains can grow significantly with both the degree of passivity and the size of the process. In the warehouse system, PSBF outperformed the alternative methods by up to 49%, which is a substantial saving considering the size of the state space (more than $10^{45}$ states). Furthermore, the computational gains where much higher in the centralised control mode than in the decentralised control mode, since the latter had a significantly lower degree of passivity. Therefore, this again shows that high degrees of passivity can bear great potential for the filtering task.

## 7. No Free Lunch for PSBF

Our view is that no belief filtering method is generally suited for all types of processes. Instead, each method assumes a certain structure in the process (explicitly or implicitly) which it attempts to exploit in order to render the filtering task more tractable. Typically, the methods are tailored in such a way with respect to this structure that they perform well if the structure is present in the process, but suffer a significant loss in performance if the structure is absent. For instance, PF works best in processes with low degrees of uncertainty, since this means that fewer state samples are needed for acceptable approximations. On the other hand, the number of samples needed for acceptable approximations can grow substantially with the degree of uncertainty in the process (as shown in our experiments). As another example, BK works best in processes with little correlation between state variables, since this means that the belief factors will be small and can be processed efficiently. However, if there are many variables which are strongly correlated, then BK typically becomes infeasible. Therefore, these structural assumptions have to be taken into account when choosing a filtering method for a specific process.

A formal account of this view is given by the "No Free Lunch" theorems (Wolpert & Macready, 1997, 1995) which state that, intuitively speaking, any two algorithms have equivalent performance when averaged over all possible instances of the problem. In other words, if there are classes of problem instances for which algorithm A has better performance than algorithm B, then there must be other classes of problem instances for which A has worse performance than B. Then, the question is: for what class of problem instances (that is, processes) can PSBF be expected to achieve good performance? This class is essentially described by the following three criteria:

**Degree of passivity** — PSBF attempts to accelerate the filtering task by omitting the transition step for as many belief factors as possible. This depends on the passivity of the variables in the state clusters. In the ideal case, the process exhibits a high degree of passivity such that PSBF can omit the transition step for many belief factors. In the worst case, the process has no passive variables at all, and PSBF has to update all belief factors in the transition step. However, as discussed in Section 5.5, a high degree of passivity is not necessarily sufficient to infer that many clusters can be skipped in the transition step, since the passive variables could be distributed in such a way that no cluster can be skipped (e.g. if the passive variables are distributed uniformly amongst the state clusters). Therefore, in an optimal case, the passivity is concentrated on correlated state variables such that passive variables end up in the same clusters.

**Size of state clusters** — The space and time complexity of the belief state representation in PSBF is exponential in the size of the largest state cluster (cf. Section 5.5). Therefore, in the ideal case, the relevant variable correlations can be captured in small state clusters and the cost of storing the belief factors and performing the update procedures is small. In the worst case, large state clusters are required to retain the variable correlations and the cost of storing and updating belief factors is large. Another reason why the state clusters should be small is because of the way in which PSBF performs the transition step. One pre-requisite for omitting the transition step for a belief factor is that all variables in the corresponding cluster are passive. If there are many variables in one cluster, then it is less likely that all variables in the cluster are passive, and, therefore, it is less likely that the cluster can be skipped.

**Structure of observations** — A third criterion, though arguably less important than the other criteria, is the structure of the observations (i.e. the way in which the observation variables depend on the state variables) and the size of the observation clusters ($\hat{C}_l$). PSBF attempts to accelerate the observation step by skipping over all those state clusters whose variables are structurally independent of the observation, and, if a cluster cannot be skipped, by incorporating only those observation clusters which are relevant to the update. Therefore, in the ideal case, only a fraction of the state clusters depend on the observation, and the relevant correlations between observation variables can be captured in small observation clusters. In the worst case, all state clusters depend on the observation in some sense, and the structure of the observation does not allow for an efficient clustering.

Thus, in summary, PSBF is most suitable for processes with high degrees of passivity and in which the relevant variable correlations can be captured in small state and observation clusters. On the other hand, PSBF may not be suitable if there is no or only low degrees of passivity, and if large state and observation clusters are necessary to retain the relevant variable correlations in the process.

In addition to identifying the class of processes for which a filtering method is suitable, it is also important to justify the practical relevance of this class. In this work, we are interested in robotic and other physical decision processes (as shown by our examples and experiments). Such systems typically exhibit a number of features: First of all, robotic systems usually have some causal structure (e.g. Mainzer, 2010; Pearl, 2000). Passivity, as a specific type of causality, can be observed in many robotic systems, including the robot arm used in our examples and the multi-robot warehouse system in Section 6.2. Furthermore, robotic systems most typically have a modular structure, in which each module is responsible for a specific subtask and may interact with other modules. This modular structure often allows for an efficient clustering, in the sense that each module corresponds to a cluster of correlated state variables. Finally, the sensors used in robotic systems typically only provide information about certain aspects of the system, and some components of the system may not benefit from some of the sensor information. In other words, there are independencies between state and observation variables. These features correspond to the criteria (above) which specify the class of processes for which PSBF is a suitable filtering method. Therefore, we believe that this class is practically justified.

## 8. Conclusion

Inferring the state of a stochastic process can be a difficult technical challenge in complex systems with large state spaces. The key to developing efficient solutions is to identify special structure in the process, e.g. in the topology and parameterisation of dynamic Bayesian networks, which can be leveraged to render the filtering task more tractable.

To this end, the present article explored the idea of automatically detecting and exploiting causal structure in order to accelerate the belief filtering task. We considered a specific type of causal relation, termed passivity, which pertains to how state variables cause changes in other state variables. To demonstrate the potential of exploiting passivity, we developed a novel filtering method, PSBF, which uses a factored belief state representation and exploits passivity to perform selective updates over the belief factors. PSBF produces exact belief states under certain assumptions and approximate belief states otherwise. We showed empirically, in synthetic processes with varying sizes and degrees of passivity as well as in an example of a complex multi-robot system, that PSBF can be faster than several alternative methods while achieving competitive accuracy. In particular, our results showed that the computational gains can grow significantly with the size of the process and the degree of passivity.

Our work demonstrates that if a system exhibits much causal structure, then there can be great potential in exploiting this structure to render the filtering task more tractable. In particular, our experiments support our initial hypothesis that factored beliefs and passivity can be a useful combination in large processes. This insight is relevant for complex processes with high degrees of causality, such as robots used in homes, offices, and industrial factories, where the filtering task may constitute a major impediment due to the often very large state space of the system.

There are several potential directions for future work. For example, it would be useful to know if the definition of passivity could be relaxed such that more variables fall under this definition, and such that the principal idea behind PSBF is still applicable. One such relaxation could be in the form of *approximate* passivity, which allows for small probabilities that passive variables change values even if the relevant parents remain unchanged. In addition, it would be interesting to know if the idea of performing selective updates over belief factors (via passivity) could also be applied to other existing methods that use a factored belief state representation (cf. Section 2.1). Finally, another useful avenue for future work would be to formulate additional types of causal relations which can be exploited in ways similar to how PSBF exploits passivity, or perhaps in ways other than that.

## Acknowledgements

This article is the result of a long debate on the presented topic, and in the process benefited from a number of discussions and suggestions. In particular, the authors wish to thank anonymous reviewers from the NIPS'12 and UAI'13 conferences as well as the Journal of AI Research; attendees of the workshop on "Advances in Causal Inference" held at UAI'15; and our colleagues in the School of Informatics at The University of Edinburgh. Furthermore, the authors acknowledge the financial support of the German National Academic Foundation, the UK Engineering and Physical Sciences Research Council (grant number EP/H012338/1), and the European Commission (TOMSY Grant Agreement 270436).

## Appendix A. Proof of Theorem 1

To prove Theorem 1, it will be useful to first establish the following lemma:

**Lemma 1.** If (A1) holds and all $x_i^{t+1} \in C_k$ are passive in $\Delta^a$, then

$$\forall s, s' : T_k^a(s, s_k') = 1 \Leftrightarrow s_k = s_k'.$$

*Proof.*

$\Rightarrow$: The fact of (A1) means that $\Phi_{a,i} \subseteq C_k$ for all $x_i^{t+1} \in C_k$. Since all $x_i^{t+1} \in C_k$ are passive in $\Delta^a$, it follows that all $x_j^t \in \Phi_{a,i}$ are passive in $\Delta^a$, for all $\Phi_{a,i}$. Therefore, given $T_k^a(s, s_k') = 1$ and clause (ii) in Definition 3, it follows that $s_k = s_k'$.

$\Leftarrow$: Follows directly by (A1) and the fact that all $x_i^{t+1} \in C_k$ are passive in $\Delta^a$. $\qquad\square$

Using Lemma 1, we can give a compact proof of Theorem 1:

**Theorem 2.** If (A1) and (A2) hold, and if all $x_i^{t+1} \in C_k$ are passive in $\Delta^{a^t}$, then

$$\forall s : \hat{b}_k^{t+1}(s_k) = b_k^t(s_k).$$

*Proof.*

$$
\begin{aligned}
\hat{b}_k^{t+1}(s_k') \;&=\; \eta_1 \sum_{\bar{s} \in S(pa_{a^t}^t(C_k))} T_k^{a^t}(\bar{s}, s_k') \prod_{k' : [\exists x_i^{t+1} \in C_{k'} \,:\, x_i^t \in pa_{a^t}^t(C_k)]} b_{k'}^t(\bar{s}_{k'}) \\[2mm]
&\overset{\text{Lem1}}{=}\; \eta_1 \sum_{\bar{s} \in S(pa_{a^t}^t(C_k)) : \bar{s}_k = s_k'} T_k^{a^t}(\bar{s}, s_k') \prod_{k' : [\exists x_i^{t+1} \in C_{k'} \,:\, x_i^t \in pa_{a^t}^t(C_k)]} b_{k'}^t(\bar{s}_{k'}) \\[2mm]
&=\; \eta_1\, b_k^t(s_k) \underbrace{\sum_{\bar{s} \in S(pa_{a^t}^t(C_k)) : \bar{s}_k = s_k'} T_k^{a^t}(\bar{s}, s_k') \prod_{k' \neq k : [\exists x_i^{t+1} \in C_{k'} \,:\, x_i^t \in pa_{a^t}^t(C_k)]} b_{k'}^t(\bar{s}_{k'})}_{\overset{(A1)}{=}\,1} \\[2mm]
&=\; \eta_1\, b_k^t(s_k) \\[2mm]
&=\; b_k^t(s_k). \quad (\eta_1 = 1 \text{ since } b_k^t \text{ normalised})
\end{aligned}
$$

$\qquad\square$

# Appendix B. Proof of Theorem 2

To prove Theorem 2, we first note the following proposition:

**Proposition 1.** If all $x_i^{t+1} \in C_k$ are marginally independent of all $y_j^{t+1} \in Y^{t+1}$ in $\Delta^{a^t}$, then

$$\forall s, s' : \left( \wedge_{k' \neq k} s_{k'} = s'_{k'} \right) \to \Omega^a(s, o^t) = \Omega^a(s', o^t).$$

This proposition follows directly by definition.

Using Proposition 1, we can give a compact proof of Theorem 2:

**Theorem 2.** If all $x_i^{t+1} \in C_k$ are marginally independent of all $y_j^{t+1} \in Y^{t+1}$ in $\Delta^{a^t}$, then

$$\forall s : b_k^{t+1}(s_k) = \hat{b}_k^{t+1}(s_k).$$

*Proof.*

$$
\begin{aligned}
b_k^{t+1}(s'_k) &= \eta_2 \, \hat{b}_k^{t+1}(s'_k) \underbrace{\sum_{\bar{s} \in S(pa_{a^t}^{t+1}(Y^{t+1})) : \bar{s}_k = s'_k} \Omega^{a^t}(\bar{s}, o^{t+1}) \prod_{k' \neq k : C_{k'} \cap pa_{a^t}^{t+1}(Y^{t+1}) \neq \emptyset} \hat{b}_{k'}^{t+1}(\bar{s}_{k'})}_{\overset{\text{Prop1}}{=} \text{constant } \alpha, \text{ independent of } s'_k} \\
&= \frac{\hat{b}_k^{t+1}(s'_k) \, \alpha}{\sum_{s''_k} \hat{b}_k^{t+1}(s''_k) \, \alpha} \\
&= \frac{\hat{b}_k^{t+1}(s'_k)}{\sum_{s''_k} \hat{b}_k^{t+1}(s''_k)} \\
&= \hat{b}_k^{t+1}(s'_k).
\end{aligned}
$$

$\square$

## Appendix C. Mixture of Gaussians

Algorithm 4 provides a simple procedure that randomly generates a mixture of Gaussians (i.e. a set of normal distributions) for the synthetic processes in Section 6.1. The algorithm takes as input the number $n$ of state variables and returns a set $\mathbb{G}$ of Gaussians whose means are in the set $\{1, ..., n\}$. The number of Gaussians, their means, and their variances are chosen automatically so as to achieve good "coverage" of state variables while minimising the (visual) overlap of Gaussians. See Figure 7 for an example.

---

**Algorithm 4** MIXTUREOFGAUSSIANS($n$)

---

1: **Input:** number of state variables $n$

2: **Parameters:** $\lambda \leftarrow 4$, $\sigma_{\min} \leftarrow \frac{5}{\lambda}$, $\sigma_{\max} \leftarrow \frac{n}{10}$

3: **Output:** mixture of Gaussians $\mathbb{G}$

4: $\mathbb{G} \leftarrow \emptyset$

5: $\mathbb{R} \leftarrow \{(1, ..., n)\}$

6: **while** $\mathbb{R} \neq \emptyset$ **do**

7: $\quad R \leftarrow$ next element of $\mathbb{R}$

8: $\quad \mathbb{R} \leftarrow \mathbb{R} \setminus \{R\}$

9: $\quad \mu \leftarrow R(\lceil \texttt{rand} * |R| \rceil)$ // $\texttt{rand}$ *returns random number from* $(0, 1)$

10: $\quad \beta \leftarrow \lambda^{-1} \min[\mu - R(1), R(|R|) - \mu]$

11: $\quad \sigma \leftarrow \min[\sigma_{\max}, \max[\sigma_{\min}, \texttt{rand} * \beta]]$

12: $\quad \mathbb{G} \leftarrow \mathbb{G} \cup \{(\mu, \sigma^2)\}$ // *mean and variance of Gaussian*

13: $\quad R_- \leftarrow (R(1), R(2), ..., R(p))$ such that $R(p) < \mu - \sigma\lambda$

14: $\quad R_+ \leftarrow (R(q), R(q+1), ..., R(|R|))$ such that $R(q) > \mu + \sigma\lambda$

15: $\quad$ **if** $R_- \neq \emptyset$ **then**

16: $\quad\quad \mathbb{R} \leftarrow \mathbb{R} \cup \{R_-\}$

17: $\quad$ **if** $R_+ \neq \emptyset$ **then**

18: $\quad\quad \mathbb{R} \leftarrow \mathbb{R} \cup \{R_+\}$

19: **return** $\mathbb{G}$

---

# References

Astrom, K. (1965). Optimal control of Markov processes with incomplete state information. *Journal of Mathematical Analysis and Applications*, *10*, 174–205.

Boutilier, C., Dean, T., & Hanks, S. (1999). Decision-theoretic planning: structural assumptions and computational leverage. *Journal of Artificial Intelligence Research*, *11*(1), 1–94.

Boutilier, C., Friedman, N., Goldszmidt, M., & Koller, D. (1996). Context-specific independence in Bayesian networks. In *Proceedings of the 12th Conference on Uncertainty in Artificial Intelligence*, pp. 115–123.

Boyen, X., & Koller, D. (1998). Tractable inference for complex stochastic processes. In *Proceedings of the 14th Conference on Uncertainty in Artificial Intelligence*, pp. 33–42.

Boyen, X., & Koller, D. (1999). Exploiting the architecture of dynamic systems. In *Proceedings of the 16th National Conference on Artificial Intelligence*, pp. 313–320.

Brafman, R. (1997). A heuristic variable grid solution method for POMDPs. In *Proceedings of the 14th National Conference on Artificial Intelligence*, pp. 727–733.

D'Andrea, R., & Wurman, P. (2008). Future challenges of coordinating hundreds of autonomous vehicles in distribution facilities. In *Proceedings of the IEEE International Conference on Technologies for Practical Robot Applications*, pp. 80–83.

Dean, T., & Kanazawa, K. (1989). A model for reasoning about persistence and causation. *Computational Intelligence*, *5*, 142–150.

Dias, M., Zlot, R., Kalra, N., & Stentz, A. (2006). Market-based multirobot coordination: a survey and analysis. *Proceedings of the IEEE*, *94*(7), 1257–1270.

Doucet, A., de Freitas, N., & Gordon, N. (2001). *Sequential Monte Carlo Methods in Practice*. Springer Science & Business Media.

Doucet, A., De Freitas, N., Murphy, K., & Russell, S. (2000). Rao-Blackwellised particle filtering for dynamic Bayesian networks. In *Proceedings of the 16th Conference on Uncertainty in Artificial Intelligence*, pp. 176–183.

Geiger, D., Verma, T., & Pearl, J. (1989). d-separation: from theorems to algorithms. In *Proceedings of the 5th Conference on Uncertainty in Artificial Intelligence*, pp. 139–148.

Gordon, N., Salmond, D., & Smith, A. (1993). Novel approach to nonlinear/non-Gaussian Bayesian state estimation. In *IEE Proceedings F (Radar and Signal Processing)*, Vol. 140, pp. 107–113.

Hart, P., Nilsson, N., & Raphael, B. (1968). A formal basis for the heuristic determination of minimum cost paths. In *IEEE Transactions on Systems Science and Cybernetics*, Vol. 4, pp. 100–107.

Hauskrecht, M. (2000). Value-function approximations for partially observable Markov decision processes. *Journal of Artificial Intelligence Research*, *13*, 33–94.

Heckerman, D. (1993). Causal independence for knowledge acquisition and inference. In *Proceedings of the 9th Conference on Uncertainty in Artificial Intelligence*, pp. 122–127.

Heckerman, D., & Breese, J. (1994). A new look at causal independence. In *Proceedings of the 10th Conference on Uncertainty in Artificial Intelligence*, pp. 286–292.

Kaelbling, L., Littman, M., & Cassandra, A. (1998). Planning and acting in partially observable stochastic domains. *Artificial intelligence*, *101*(1), 99–134.

Koller, D., & Friedman, N. (2009). *Probabilistic Graphical Models: Principles and Techniques*. The MIT Press.

Kullback, S., & Leibler, R. (1951). On information and sufficiency. *The Annals of Mathematical Statistics*, *22*(1), 79–86.

Lauritzen, S., & Spiegelhalter, D. (1988). Local computations with probabilities on graphical structures and their application to expert systems. *Journal of the Royal Statistical Society. Series B (Methodological)*, *50*(2), 157–224.

Lovejoy, W. (1991). Computationally feasible bounds for partially observed Markov decision processes. *Operations Research*, *39*, 162–175.

Mainzer, K. (2010). Causality in natural, technical, and social systems. *European Review*, *18*, 433–454.

Minka, T., Winn, J., Guiver, J., & Knowles, D. (2012). Infer.NET 2.5.. Microsoft Research Cambridge. http://research.microsoft.com/infernet.

Murphy, K. (2001). The Bayes net toolbox for Matlab. *Computing Science and Statistics*, *33*(2), 1024–1034. https://code.google.com/p/bnt/.

Murphy, K., & Weiss, Y. (2001). The factored frontier algorithm for approximate inference in DBNs. In *Proceedings of the 17th Conference on Uncertainty in Artificial Intelligence*, pp. 378–385.

Murphy, K. (2002). *Dynamic Bayesian Networks: Representation, Inference and Learning*. Ph.D. thesis, University of California, Berkeley.

Ng, B., Peshkin, L., & Pfeffer, A. (2002). Factored particles for scalable monitoring. In *Proceedings of the 18th Conference on Uncertainty in Artificial Intelligence*, pp. 370–377.

Pasula, H., Zettlemoyer, L., & Kaelbling, L. (2007). Learning symbolic models of stochastic domains. *Journal of Artificial Intelligence Research*, *29*, 309–352.

Pearl, J. (1988). *Probabilistic Reasoning in Intelligent Systems: Networks of Plausible Inference*. Morgan Kaufmann.

Pearl, J. (2000). *Causality: Models, Reasoning, and Inference*. Cambridge University Press.

Pineau, J., Gordon, G., & Thrun, S. (2003). Point-based value iteration: an anytime algorithm for POMDPs. In *Proceedings of the 18th International Joint Conference on Artificial Intelligence*, Vol. 18, pp. 1025–1032.

Poole, D., & Zhang, N. (2003). Exploiting contextual independence in probabilistic inference. *Journal of Artificial Intelligence Research*, *18*, 263–313.

Poupart, P., & Boutilier, C. (2000). Value-directed belief state approximation for POMDPs. In *Proceedings of the 16th Conference on Uncertainty in Artificial Intelligence*, pp. 497–506.

Poupart, P., & Boutilier, C. (2001). Vector-space analysis of belief-state approximation for POMDPs. In *Proceedings of the 17th Conference on Uncertainty in Artificial Intelligence*, pp. 445–452.

Poupart, P., & Boutilier, C. (2002). Value-directed compression of POMDPs. In *Advances in Neural Information Processing Systems*, pp. 1547–1554.

Roy, N., Gordon, G., & Thrun, S. (2005). Finding approximate POMDP solutions through belief compression. *Journal of Artificial Intelligence Research*, *23*, 1–40.

Smith, T., & Simmons, R. (2005). Point-based POMDP algorithms: improved analysis and implementation. In *Proceedings of the 21st Conference on Uncertainty in Artificial Intelligence*, pp. 542–549.

Sondik, E. (1971). *The Optimal Control of Partially Observable Markov Processes*. Ph.D. thesis, Stanford University.

Srinivas, S. (1993). A generalization of noisy-or model. In *Proceedings of the 9th Conference on Uncertainty in Artificial Intelligence*, pp. 208–215.

Washington, R. (1997). BI-POMDP: bounded, incremental partially-observable Markov-model planning. In *Recent Advances in AI Planning*, pp. 440–451. Springer.

Wolpert, D., & Macready, W. (1995). No free lunch theorems for search. Tech. rep. SFI-TR-95-02-010, Santa Fe Institute.

Wolpert, D., & Macready, W. (1997). No free lunch theorems for optimization. *IEEE Transactions on Evolutionary Computation*, *1*(1), 67–82.

Wurman, P., D'Andrea, R., & Mountz, M. (2008). Coordinating hundreds of cooperative, autonomous vehicles in warehouses. *AI Magazine*, *29*(1), 9.

Zhang, N., & Poole, D. (1996). Exploiting causal independence in Bayesian network inference. *Journal of Artificial Intelligence Research*, *5*, 301–328.

Zhou, R., & Hansen, E. (2001). An improved grid-based approximation algorithm for POMDPs. In *Proceedings of the 17th International Joint Conference on Artificial Intelligence*, pp. 707–716.